\documentclass[lettersize,journal]{IEEEtran}
\usepackage{amsmath,amsfonts}
\usepackage{algorithmic}
\usepackage{algorithm}
\usepackage{array}
\usepackage[caption=false,font=footnotesize,labelfont=rm,textfont=rm]{subfig}
\usepackage{textcomp}
\usepackage{stfloats}
\usepackage{url}
\usepackage{verbatim}
\usepackage{graphicx}
\usepackage{cite}
\usepackage{pifont}
\usepackage{booktabs}
\usepackage{multirow}
\usepackage{xcolor}
\usepackage{threeparttable}
\usepackage{ulem}

\begin{document}
	
	\title{CFVNet: An End-to-End Cancelable Finger Vein Network for Recognition}
	
	\author{Yifan Wang,~\IEEEmembership{Student Member,~IEEE,} Jie Gui,~\IEEEmembership{Senior Member,~IEEE,} Yuan Yan Tang,~\IEEEmembership{Life Fellow,~IEEE,} and James Tin-Yau Kwok,~\IEEEmembership{Fellow,~IEEE,}
		\thanks{This work was supported in part by the grant of the National Science Foundation of China under Grant 62172090; Start-up Research Fund of Southeast University under Grant RF1028623097. We thank the Big Data Computing Center of Southeast University for providing the facility support on the numerical calculations. (Corresponding author: Jie Gui.)}
		\thanks{Y. Wang and J. Gui are with the School of Cyber Science and Engineering, Southeast
			University, Nanjing 210000, China. J. Gui is also with Engineering Research Center of Blockchain Application, Supervision and Management (Southeast University), Ministry of Education; Purple Mountain Laboratories, Nanjing 210000, China (e-mail: 230239767@seu.edu.cn, guijie@seu.edu.cn).}
		\thanks{Yuan Yan Tang is with the Department of Computer and Information Science, University of Macao, Macao 999078, China (e-mail: yytang@um.edu.mo).}
		\thanks{James Tin-Yau Kwok is with the Department of Computer Science and
			Engineering, The Hong Kong University of Science and Technology, Hong
			Kong 999077, China (e-mail: jamesk@cse.ust.hk).}
		\thanks{This manuscript has been accepted by IEEE Transactions on Information Forensics \& Security, DOI: 10.1109/TIFS.2024.3436528}
	}
	\markboth{Journal of \LaTeX\ Class Files,~Vol.~14, No.~8, August~2021}%
	{Shell \MakeLowercase{\textit{et al.}}: A Sample Article Using IEEEtran.cls for IEEE Journals}
	
	\IEEEpubid{0000--0000/00\$00.00~\copyright~2021 IEEE}
	
	\maketitle
	
	\begin{abstract}
		Finger vein recognition technology has become one of the primary solutions for
		high-security identification systems. However, it still has information leakage
		problems, which seriously jeopardizes user's privacy and anonymity and cause great
		security risks. In addition, there is no work to consider a fully integrated secure finger vein recognition system. So, different from the previous systems, we integrate preprocessing
		and template protection into an integrated deep learning model.
		We propose an end-to-end cancelable finger vein network
		(CFVNet), which can be used to design an secure finger vein recognition system. It includes a plug-and-play
		BWR-ROIAlign unit, which consists of three sub-modules: Localization, Compression
		and Transformation. The localization module achieves automated localization of
		stable and unique finger vein ROI. The compression module losslessly removes
		spatial and channel redundancies. The transformation module uses the proposed BWR
		method to introduce unlinkability, irreversibility and revocability to the system. BWR-ROIAlign can directly plug into the model to introduce the above features for DCNN-based finger vein recognition systems. We perform extensive experiments on four public datasets to
		study the performance and cancelable biometric attributes of the CFVNet-based recognition system. The
		average accuracy, EERs and $D_\leftrightarrow^{sys}$ on the four datasets 
		are 99.82\%, 0.01\% and 0.025, respectively, and achieves competitive
		performance compared with the state-of-the-arts.
	\end{abstract}
	
	
	\section{Introduction} \label{sec:intro}
	\IEEEPARstart{W}{ith} the rapid development in the information society, traditional
	identification technologies based on password or ID card can no longer meet the
	security demand. In recent years, applications of biometrics in finance,
	security, medical and other fields have shown many advantages
	\cite{shaheed2022recent}. Biometrics allows
	intelligent machines to automatically capture, process, analyse and recognize 
	digital physiological or behavioral characteristic signals of the human body,
	and is at the forefront of artificial intelligence development. The
	main biometric features include face \cite{yang2019learning}, fingerprint
	\cite{10100723}, veins \cite{ri2022robust} and gait \cite{10242019}. 
	As a
	representative technology for second-generation biometrics, 
	finger vein recognition (FVR)
	has become the
	mainstream in recent research because of its convenience and higher security
	quality. It
	can be divided into several main steps: image acquisition, preprocessing, feature extraction and
	recognition. Deep learning based FVR systems can 
	automatically complete the extraction and recognition steps. On the other hand,
	preprocessing, while being a key
	step in FVR 
	\cite{wang2023residual}, 
	requires the design of special processing
	operations (such as region-of-interest extraction, image enhancement, and feature
	alignment)
	to suit the different acquisition devices and application scenarios. 
	It can seriously affect the
	convenience, integrality, and automation level of the recognition system.
	
	With biometrics being widely used, the information leakage problem can seriously
	jeopardize users' privacy and anonymity \cite{patel2015cancelable}. Although
	biometrics is more difficult to be fraudulently copied or forged than traditional
	identification techniques, 
	attackers can still track 
	activities 
	of subjects
	in different domains 
	enrolled 
	through biometrics
	\cite{kauba2022towards}. Once a biometric feature is compromised, the 
	system's security will decrease drastically. This biometric feature can no longer
	be used,
	further limiting the number of biometric features that can be applied
	on  a subject. As biometric information cannot be revoked
	and redistributed, 
	Ratha et al. \cite{ratha2001enhancing}
	proposed to solve this problem  by
	introducing a concept called cancelable biometrics (CB), which is defined as a
	biometric signal that is subjected to an intentional and repeatable distortion
	transformation operation. It allows the biometric templates to be authenticated in
	the protected domain to achieve biometrics protection. 
	According to the
	ISO/IEC 24745 standard on biometric information protection \cite{iso2022}, a
	properly cancelable biometric should have the following attributes.
	\IEEEpubidadjcol
	\begin{itemize}
		
		\item \textit{\textbf{Irreversibility}}: Given a protected template, it should not be possible to reconstruct the original biometric sample.
		
		\item \textit{\textbf{Revocability}}: From a given biometric sample, it should be possible to issue multiple protected templates.
		
		\item \textit{\textbf{Unlinkability}}: Given two protected templates that are
		generated by the same instances and stored in different systems, it should not
		be feasible to determine if they belong to the same subject.
		
		\item \textit{\textbf{Performance}}: The system's recognition performance
		should not be significantly degraded when using the 
		biometric template protection, and should not be sensitive to the parameters of the template protection step employed.
		
	\end{itemize}
	
	Unlike traditional recognition methods, FVR systems based on deep convolutional neural
	network (DCNN) do not need to save the feature templates into a database, and thus
	there is no risk that the templates will be stolen \cite{9530463}. However, an
	attacker can hack into the finger vein system to analyze the finger vein
	information output by the network. A main component of the DCNN is the
	convolutional operation, and one can recover the original vein template 
	to some extent 
	via deconvolution 
	\cite{ren2021finger}. This 
	can become a security risk 
	for the DCNN-based FVR system 
	once the template
	is recovered. To solve these problems, researchers combine biometric template protection with DCNN-based FVR system to design more secure
	systems. Using simple and effective irreversible transformations to reorganize or
	distort the biometric features, the CB template protection is
	designed to make it impossible for the attacker to restore the original features
	using the protected template. However, the template protection scheme proposed
	in previous works \cite{ren2021finger} requires first transforming the raw feature templates, and then they can be used as inputs to a DCNN-based recognition system. Similar to the preprocessing task, the template protection task is also separated from the DCNN model, which makes it impossible to realize a DCNN-based end-to-end recognition system.
	
	To solve the above problems and achieve integrated finger vein recognition, we propose an end-to-end cancelable finger vein network CFVNet for efficient and secure recognition system. By directly
	inputting the finger vein images obtained from the collection device into the
	recognition system, we integrate preprocessing, feature extraction, template
	protection and recognition into a complete deep learning model. 
	
	The proposed
	system has a 
	block warping remapping ROI align (BWR-ROIAlign)
	plug-and-play module, which consists of three
	sub-modules: localization, compression, and transformation. 
	The localization sub-module localizes stable and unique regions in the same instances.
	This is important because finger vein feature maps extracted by a
	standard DCNN are sparse. For feature transformation-based biometric template
	protection, slight translation or rotation of the same instances can show obvious
	intra-class differences in the transformed feature template. 
	The 
	stable regions extracted by the 
	localization sub-module 
	allow feature alignment to effectively narrow intra-class differences. 
	This approach can be regarded as an effective solution to automated finger vein image ROI localization. 
	
	For the compression sub-module, firstly, the spatial redundancy in the vein
	images is removed according to the ROI coordinates, and the unique stable finger
	vein ROI is retained. As the finger vein pattern is relatively simple,
	a standard DCNN may then produce redundant feature channels which increases the computational burden.
	As the key information may only exist in a small number of feature channels, the
	compression sub-module removes redundant spatial regions and feature channels by linear weighted fusion. 
	
	Finally, the transformation sub-module re-divides the compact feature maps into
	same-sized blocks in the source domain, and then randomly samples part of the
	blocks, maps them to the new feature maps in random order, and
	uses mesh warping to dilute the response connection between the boundary pixels in
	the blocks. The transformation sub-module completes the conversion of source
	domain features to protected domain features, extract and analyze the protected
	domain features, and finally use the transformed features to complete the
	identification in the protected domain. This introduces unlinkability,
	irreversibility and revocability with higher security to the identification system. 
	
	The major contributions of this paper are as follows.
	
	\begin{itemize}
		\item We propose an end-to-end cancelable finger vein network (CFVNet) for the FVR system. To the best of our knowledge, this is the first study that integrates finger vein image preprocessing, template protection, and recognition tasks into a complete DCNN model. 
		
		\item We design a novel block warping remapping ROI align (BWR-ROIAlign)
		module, which consists of three sub-modules: localization, compression and
		transformation. The localization module automatically locates and extracts
		stable and unique finger vein ROIs. The compression module improves 
		efficiency of the recognition system by reducing spatial and channel
		redundancies. The transformation module introduces unlinkability and
		irreversibility to the biometric recognition system. The BWR-ROIAlign module is
		a plug-and-play architectural unit.
		
		\item 
		This is the first work to label finger vein ROIs. We label ROIs in four
		publicly available datasets, and conduct comprehensive experiments on these
		datasets to verify the recognition and localization performance of the proposed model. We also conduct analysis on unlinkability, revocability, and irreversibility of the CFVNet-based FVR system.
		
	\end{itemize}
	
	The rest of the paper is organized as follows. Section~\ref{sec:related} briefly
	reviews the related works. Section \ref{sec:method}
	introduces the proposed CFVNet for FVR.
	Section 
	\ref{sec:expt}
	presents the experimental setup, results, and analysis. Finally, the
	paper concludes with Section
	\ref{sec:conclusion}.

	\section{Related Work}
	\label{sec:related}
	In this section, we briefly review the work related to FVR and cancelable finger vein biometrics.
	
	\subsection{Finger Vein Recognition}
	FVR has four main steps: image acquisition, preprocessing, feature extraction and
	recognition. In the past decade, 
	many effective solutions 
	have been proposed 
	for the recognition task. These methods can be categorized as using manual
	features \cite{ri2022robust} or deep learning \cite{10023509}. In early FVR research, 
	vein features 
	(mainly based on the vein pattern, orientation or details)
	are usually produced
	manually 
	by feature engineering. 
	Miura et al. proposed the maximum curvature \cite{miura2007extraction} and
	repeated line tracking \cite{miura2004feature} to manually extract vein
	features. Some common gradient operators,
	such as LBP, LDP, ELBP, Gabor filter \cite{ri2022robust} and SIFT, 
	are also used for vein feature extraction.
	Methods such as PCA, LDA, and Dictionary Learning \cite{yang2023small}
	perform spatial projection and then map the high-dimensional images to low-dimensional vein features. 
	
	Recently, the outstanding performance of DCNN in various image processing tasks has attracted extensive attention from FVR researchers. Many 
	recognition schemes 
	have been proposed.
	Razdi et al. \cite{radzi2016finger} first introduce
	the DCNN into FVR and designed a four-layer DCNN. Subsequently, researchers have
	proposed solutions to the problems associated with finger vein
	images, such as poor image quality, finger displacement/rotation, and complex
	background \cite{wang2023residual}. These include model structure design \cite{10023509}, data augmentation \cite{wang2019learning}, attention mechanism \cite{qin2022local}, loss function \cite{hou2021arcvein}, 
	learnable feature descriptors \cite{li2022learning}
	and 
	other more advanced models \cite{10201897}. These works can
	effectively improve performance of the FVR system, but they all require some
	preprocessing. For example, one has to use ROI extraction to locate a target region with stable and unique vein pattern features.
	
	A detailed review on finger vein ROI extraction 
	has been studied
	in our previous work\cite{wangroi}. Finger vein ROI extraction methods can be categorized into four types: (i) fixed
	window-based methods, (ii) thresholding-based methods, (iii) gradient operator
	based methods, and (iv) fine edge detection based methods. All these methods have drawbacks. For fixed window-based methods, the window size is
	usually fixed, and cannot adapt to changes. For thresholding and gradient
	operator based methods, the pixel values of the finger region have to be
	higher or lower than those in the background. However, in practice, 
	the finger region pixel values can have a similar distribution as the background
	region
	due to lighting conditions and/or devices.
	Moreover, many operators are very sensitive to noise. As different
	devices acquire background region images with large differences in the noise
	distribution, it is easy to take the noise as the maximum response and detect the
	wrong edges, causing poor robustness. The fine edge detection methods are mainly
	customized for different devices and application scenarios, and cannot be easily 
	applied to the other devices. The above problems lead to a separation of
	preprocessing tasks from the other tasks in deep learning based recognition systems,
	and affects the design and implementation of an integrated FVR
	framework.
	
	\subsection{Cancelable Finger Vein Biometrics}
	Compared to the other FVR tasks, research on template protection is still relatively
	scarce. Existing template protection techniques can be mainly categorized into two
	types\cite{jain2005biometric}: biometric cryptosystem and feature
	transformation (cancelable transformation) based methods. Biometric cryptosystem use biometric templates to protect the key or generate the
	key directly from biometric features. Depending on how the auxiliary data is
	obtained, these methods can be further categorized as key-binding or key-generation.
	Key-binding methods
	use a biometric-independent external key-binding biometric to obtain auxiliary
	data; while 
	key-generation methods
	generate
	the auxiliary data 
	from a biometric
	template and the key is produced directly from the auxiliary data and 
	biometric template. 
	In other words, biometric cryptosystem uses the finger vein image in the source domain, which
	makes the system vulnerable to linkage attacks. Once the finger
	vein feature template is compromised, security of the system decreases
	drastically. This feature template can no longer be used, further
	limiting availability of the biometric features.
	
	For feature transformation based methods, they generate the protected template
	$T_P$ by transforming the original template $T$ into another domain by a
	random key $K$ and transformation function $F$. Only the transformed template is
	used for recognition. When the template is compromised, one only needs to
	change the key and regenerate the protected template. Feature transformation based
	methods are categorized as reversible and irreversible, depending on whether the
	transformation process is reversible or not. The reversible
	transformation is also known as saltting or bio-hashing, which is performed by
	defining an orthogonal random transformation function using a key, through which
	the bio-features are transformed and further binarized sequences are matched. Shao
	et al. \cite{shao2022template} perform randomization and chaotic scrambling of
	finger vein features to disrupt the positional distribution of pixels, and
	binarize the chaotic features to get the protected feature vectors. Ren
	et al. \cite{ren2021finger} encrypte finger vein images using RSA and
	use the encrypted images to train a CNN model for recognition. Liu
	et al. \cite{liu2018finger} use random projection to transform the finger vein
	detail point features to get the protected template, and use a deep belief network
	as discriminator to design the recognition system. Simon et al.
	\cite{kirchgasser2020finger} propose a template protection scheme without
	alignment. They design an efficient binary feature vector representation and utilize
	IoM hashing to obtain the protected feature vector. Shahreza et al.
	\cite{shahreza2021towards} apply bio-hash to generate protected templates from
	depth features of a deep learning model.
	Non-reversible transformation
	is first proposed and applied to fingerprint template protection by
	Ratha \cite{ratha2001enhancing}.  It
	uses a parametric adjustable invertible transformation of
	biometric features to obtain a protected template which
	cannot be recovered from the original template. Kauba et al.
	\cite{kauba2022towards} test three transformation methods for
	generating revocable templates and 
	study the impact of the different methods on recognition performance,
	renewability, unlinkability and irreversibility. Yang et al.
	\cite{yang2019securing} use a binary decision diagram (BDD) to create irreversible
	templates, and utilize the feature templates in conjunction with an ML-ELM to
	construct a revocable irreversible FVR system.
	Debiasi et al.
	\cite{debiasi2019biometric} use an external key to change the gray value of an
	image, and achieve cancelable irreversible template generation in the image source
	domain.
	\begin{figure*}[t]
		\centering
		\includegraphics[scale=1,width=0.9\textwidth]{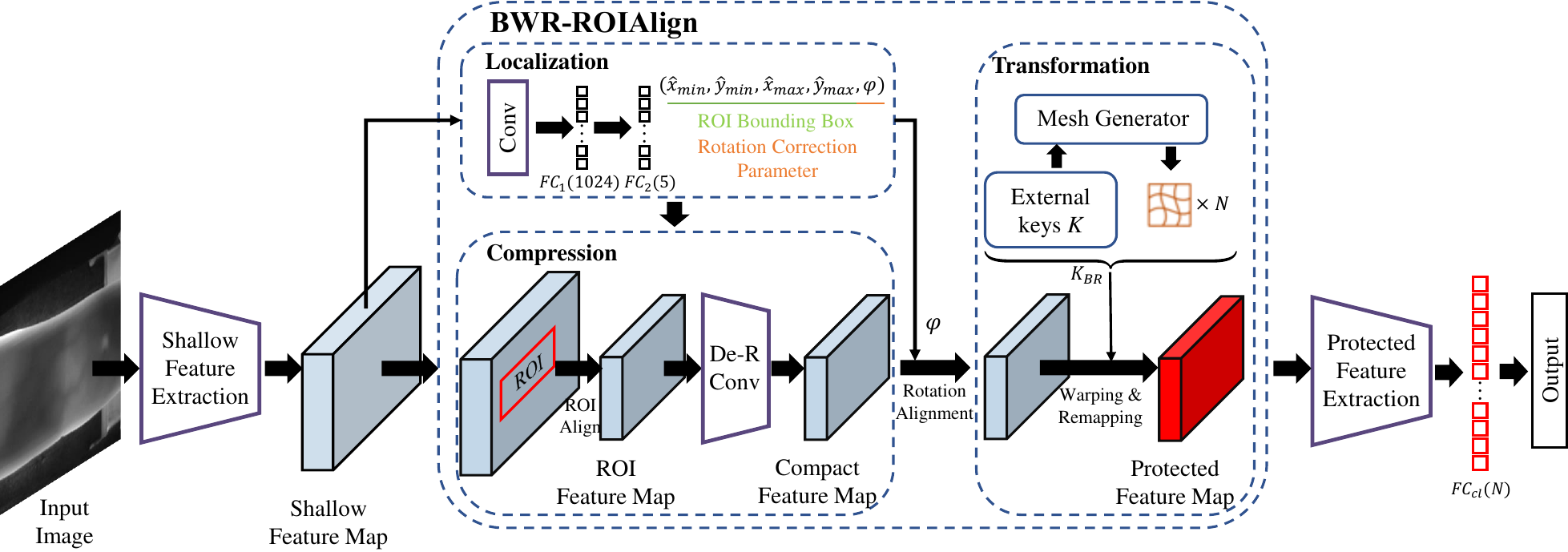}
		\caption{\centering{Cancelable finger vein network (CFVNet) overall architecture. Localization, Compression and Transformation represent the three sub-functional modules of the BWR-ROIAlign. De-R Conv is the de-redundant convolution. External key $K$ is the pseudo-random token generated during user registration.}}
		\label{fig1}
	\end{figure*}
	
	Unlike previous FVR systems, in this paper, we integrate preprocessing and template
	protection into a deep learning model. The vein feature information during the
	recognition process is highly irreversible, unlinkable and cancelable. This
	improves FVR system's cancelable biometric attributes.
	Moreover, the proposed BWR-ROIAlign is a plug-and-play module 
	for DCNN-based FVR systems, and thus provides a novel approach
	for secure and end-to-end design of FVR systems.
	
	\section{Methodology}
	\label{sec:method}
	As mentioned in Section~\ref{sec:intro}, existing works do not consider integrating individual tasks
	in FVR into a deep learning model. In order to improve the integrity of a FVR
	system, we propose a novel cancelable finger vein network for FVR. Fig.
	\ref{fig1} shows the overall architecture. In this section, we introduce the
	proposed model in detail. First, we introduce the overall design and
	functionality of the proposed CFVNet. Second, we introduce the proposed
	BWR-ROIAlign module. Finally, we introduce details of the three
	sub-modules of BWR-ROIAlign.
	
	\subsection{Cancelable Finger Vein Detector Architecture}
	Existing cancelable finger vein feature recognition systems perform
	template protection and recognition
	separately. 
	Usually, the original templates are first converted to the protection domain and
	then stored to a database, which are subsequently transmitted or put into the
	classifier for recognition. The original feature templates often need to be
	designed with fine pre-processing, including ROI extraction and feature
	alignment. This makes the cancelable FVR system cumbersome.
	
	The proposed CFVNet is 
	shown in Fig. \ref{fig1}. Its
	skeleton
	is the ResNet50, which has been commonly used in classification tasks.
	The
	model is divided into two parts based on processing vein features
	from different domains: shallow feature (source domain) extraction, and deep
	feature (protected domain) extraction. The
	shallow feature extraction includes a convolutional layer (with $7\times7$
	kernels, stride $2$, padding $3$ and $64$ channels), a batch normalization
	layer, and then ReLU activation, a max-pooling layer (with $3\times3$ kernels,
	stride $2$, and padding $1$). The protected feature extraction sub-module has the
	same structure as layers 1 to 4 in the ResNet50,
	and has an average pooling layer with $2\times2$ kernels. 
	The numbers of
	channels in these layers
	are $24$, $48$, $72$ and $96$, respectively. In the final classification layer, the protected features
	are fed to a fully-connected layer. Its output is normalized by softmax that transforms it to a probability distribution for each class.
	
	Similar to the popular two-factor biometric recognition
	system \cite{jin2004biohashing}, each user has a key which represents an
	unique tokenized pseudo-random feature transformation. This key is then used
	in the transformation sub-module to apply an unique template protection transformation. The source domain features are converted to the 
	protected feature map
	in the protected domain.
	Finally, the extracted protected
	feature vector is fed into the fully connected layer for recognition.
	
	\subsection{Block Warping Remapping ROI Align Module}
	The BWR-ROIAlign module includes three sub-modules: localization, compression
	and transformation. The localization module performs automated localization of stable
	and unique finger vein ROIs and predicts the rotational correction parameters. The compression module
	performs feature alignment and improves efficiency of the recognition system by
	reducing channel redundancy. The transformation module introduces unlinkability,
	irreversibility and revocability to the FVR system. The proposed
	BWR-ROIAlign is a plug-and-play architectural unit.
	The detailed implementation of each sub-module
	is discussed in the following.
	
	\subsubsection{\textbf{Localization}}
	The finger vein feature maps obtained by DCNN extraction have strong responses in
	key discriminative regions, but 
	weak
	in the other regions.
	This leads to 
	sparse feature maps. Finger displacement between the same instances increases the
	template intra-class difference after feature transformation. Correct feature
	alignment is an effective way to reduce the intra-class difference. Existing
	deep learning based FVR systems usually use images after preprocessing (such as ROI
	extraction and image enhancement) as inputs to the model to improve
	recognition performance. In fact, the location of unique stable
	finger vein pattern regions is performing
	feature alignment. In general, finger vein image ROI extraction focuses
	on obtaining a segmentation reference line in the horizontal and vertical
	directions, which are determined by the minimum internal tangent of the finger edge and 
	position of the finger joint cavity, respectively. Existing
	techniques \cite{wangroi} usually use
	gradient operators or threshold segmentation methods
	to acquire the finger edge or joint cavity position. However, 
	these methods 
	are not robust
	to different devices or complex noise.
	
	\begin{figure}[t]
		\centering
		\includegraphics[scale=1,width=0.47\textwidth]{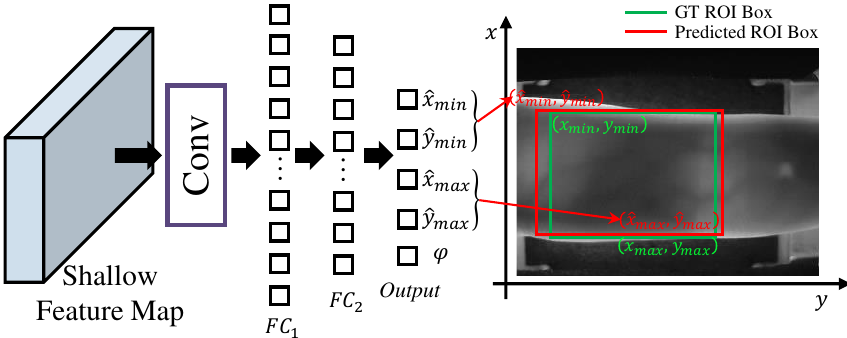}
		\caption{\centering{ROI localization sub-module.}}
		\label{fig2}
	\end{figure}
	
	Inspired by object detection and spatial transformer \cite{jaderberg2015spatial}
	techniques, we design a localization module 
	(Fig. \ref{fig2}), which consists of a convolutional
	layer (with $1\times1$ kernels, stride $2$ and $4$ channels), and two
	fully-connected layers.
	This localization module is used to predict five parameters:
	four for the ROI coordinates and one for the rotation correction parameters $\varphi$.
	ROI coordinates localization is different from that in ordinary object
	detection tasks in that we can determine the number of ROIs. Therefore, the
	localization process does not require search discrimination of the target
	region. For object detection, bounding box regression finds a mapping $f$ 
	so that 
	the offset between
	$(\hat{P}_x, \hat{P}_y, \hat{P}_w, \hat{P}_h) =f(P_x, P_y, P_w, P_h)$
	and
	the real bounding box $(G_x, G_y, G_w, G_h)$
	is reduced.
	For a set with $N$ training
	pairs ${(P^i,G^i )}_{i=1,\dots,N}$, the loss is
	\begin{equation}\label{eq1}
		Loss=\sum\nolimits_{i=1}^n{(t_*^i-{\hat{\omega}_*}^T\cdot F^i)^2},
	\end{equation}
	where
	$t$ is the offset between the regression target and the proposal, $*$
	denotes the four mappings (translation transformations $\varDelta x, \varDelta
	y$, and scale transformations $S_w, S_h$), and $F^i$ is the input feature
	map. 
	To learn the various mappings, we
	minimize the following regularized least squares objective
	w.r.t. the learning parameter $\omega$:
	\begin{equation}\label{eq2}
		W_*=\arg\min\limits_{\omega_*}\sum\nolimits_{i=1}^n{(t_*^i-{\hat{\omega}_*}^T\cdot
			F^i)^2}+\lambda {\Vert\hat{\omega}_* \Vert}^2,
	\end{equation}
	where $\lambda$ is a regularization parameter.

	General object detection tasks use the above method to realize regression to
	the bounding box, and consider the sensitivity of small objects to the coordinate
	offset. 
	The MS COCO dataset uses the absolute size, and
	defines small objects as those
	smaller than $32 \times 32$ pixels. Empirically, we find that the finger vein image ROIs
	are not 
	small objects. Hence, we do not need to learn the four 
	bounding box mappings.
	Instead, we can directly predict the real
	coordinates. The loss objective in the localization sub-module 
	is defined as
	\begin{equation}\label{eq3}
		Loss_{roi} \!=\! \begin{cases}
			\frac{1}{2}\sum_{i=1}^{N}(C^i - \hat{\omega}_*^T \cdot F^i)^2 & \text{if } |C^i - \hat{\omega}_*^T \cdot F^i| < 1, \\
			\sum_{i=1}^{N}|C^i - \hat{\omega}_*^T \cdot F^i| - \frac{1}{2} & \text{otherwise},
		\end{cases}
	\end{equation}
	where $C$ is the true finger vein ROI coordinates $(x_{min}, y_{min},
	x_{max}, y_{max} )$, 
	and ${\hat{\omega}_*}^T\cdot F^i$ is the predicted ROI coordinates $({\hat{x}_{min}},
	{\hat{y}_{min}}, {\hat{x}_{max}}, {\hat{y}_{max}}) $ obtained from the feature
	map. Parameter $\omega$ is learned by minimizing the regularized least-squares
	objective.
	
	For rotation correction, assume that the cross section of finger approximately
	resembles a circle and that the captured vein is close to the finger surface
	\cite{prommegger2021rotation}. We can use the following transformation \cite{prommegger2019extent} 
	for rotation correction:
	\begin{equation}
		\label{eq4}
		\begin{bmatrix}
			x' \\
			y' \\
		\end{bmatrix}
		=
		\begin{bmatrix}
			\cos(\varphi) & -\sin(\varphi) \\
			\sin(\varphi) & \cos(\varphi) \\
		\end{bmatrix}
		\begin{bmatrix}
			x \\
			y \\
		\end{bmatrix},
	\end{equation}
	where $(x, y)$ is the coordinates of the input feature map, and $(x^\prime,
	y^\prime)$ is the coordinates in the corrected feature map.
	Note that rotation correction is performed after compression (Section
	\ref{CM}), and the localization module only predicts the rotation correction parameter $\varphi$.

	More specifically, when a shallow feature map
	$F$ of size $(B, C, W, H)$ is used as input, the localization sub-module
	analyses
	the shallow feature information to get the ROI
	coordinates (of size $(B,4)$) and rotation correction parameters (of size
	$(B,1)$). The ROI coordinate is used in the compression module to remove
	spatial redundancy. The rotation correction parameter is used after the compression module to correct finger rotation.
	
	\subsubsection{\textbf{Compression}}
	\label{CM}
	Finger vein images have strong responses in critical discriminative regions but
	weak responses in the other regions, resulting in 
	sparse
	feature maps. Moreover, for simple finger vein patterns, redundant feature
	channels increase the computational burden. Most feature channels include no or
	only a small amount of discriminative information, and the critical discriminative
	information exists only in a few feature channels. Therefore, we design the compression sub-module 
	to remove spatial redundancy and channel redundancy.
	
	\paragraph{Spatial redundancy removal}
	Finger vein pattern features are unique, but there are still unstable features 
	(such as finger shape, position, device background, and environmental noise) 
	in the
	images acquired by unconstrained acquisition devices.
	We combine the ROI coordinates $({\hat{x}_{min}},
	{\hat{y}_{min}}, {\hat{x}_{max}}, {\hat{y}_{max}}) $ obtained from the
	localization sub-module to extract dense and stable finger vein pattern
	feature regions on the shallow feature map, and unify them into a
	$H\times W$
	feature map 
	(in the experiments, $H=32$, $W=64$).
	
	\begin{figure}[]
		\centering
		\subfloat[{\centering}][\label{fig:a1}]{
			\includegraphics[width=0.35\linewidth]{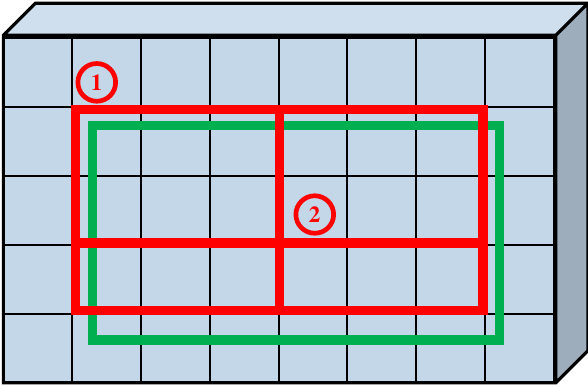}}
		\subfloat[\label{fig:b1}]{
			\includegraphics[width=0.35\linewidth]{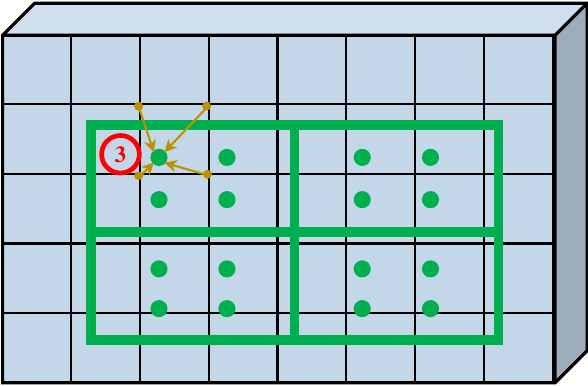}}
		\caption{\centering{Spatial redundancy removal without quantization loss.\ding{172} ROI coordinates mapped to corresponding grid cells on the feature map, \ding{173} divide the mapped grid cells into subregions of $H\times W$ number, \ding{174} Using linear interpolation in coordinate mapping and meshing.}}
		\label{fig3} 
	\end{figure}
	
	To extract specific regions in the feature map, a simpler approach is to map the
	ROI rectangular window coordinates to the corresponding grid cells on the feature
	map (Fig. \ref{fig3}-a \ding{172} operation). 
	The mapped grid cells
	are then divided
	into $H \times W$ 
	sub-regions (Fig. \ref{fig3}-a \ding{173} operation).
	Finally,
	these sub-regions 
	are sampled 
	to obtain an uniformly-sized ROI feature map.
	However, 
	non-vein regions in the finger vein feature map have low response, and so slight
	displacements in the sparse vein features can result in row or column
	variations. In particular, for block-based feature template transform methods,
	displacement variations can result in large intra-class differences. Therefore,
	proper template alignment is key to improving recognition performance of
	the protected system. We use linear interpolation (Fig. \ref{fig3}-b
	\ding{174} operation) to avoid any quantization in coordinate mapping and meshing,
	and obtain properly-aligned ROI features for subsequent sampling.
	
	\begin{figure}[]
		\centering
		\includegraphics[scale=1,width=0.3\textwidth]{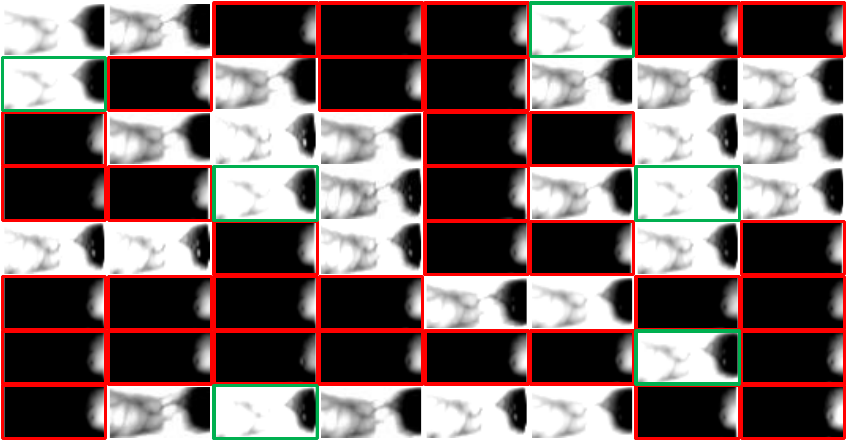}
		\caption{\centering{Redundant feature map visualization. The green boxes
				contain little feature information, while the red boxes contain almost no feature information.}}
		\label{fig4}
	\end{figure}
	\paragraph{Channel redundancy removal}
	Redundant feature channels waste computational resources. Simple
	linear operations can reduce these redundant features. 
	Fig. \ref{fig4}
	visualizes the shallow feature
	maps after removing spatial redundancy.
	Some feature maps contain only a small amount of information, and some
	do not even have any. Simple linear fusion can
	effectively remove such redundant channels.

\begin{figure}[t]
		\centering
		\includegraphics[scale=1,width=0.45\textwidth]{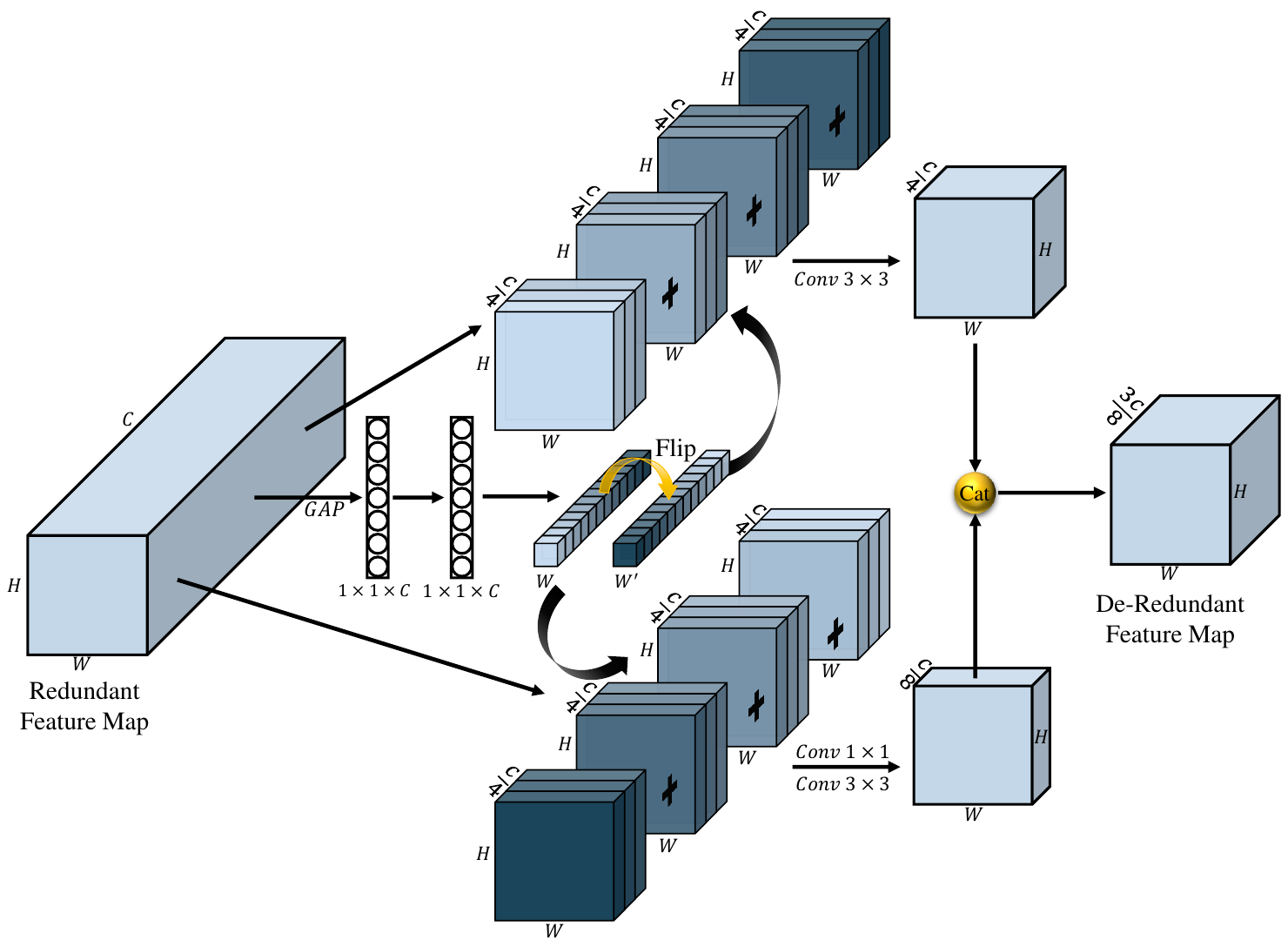}
		\caption{\centering{De-R Conv Architecture. Here,
				$\mathbf{\textbf{\textbf{\textbf{+}}}}$  represents the positions summed.}}
		\label{fig5}
	\end{figure}

\begin{figure*}[t]
		\centering
		\includegraphics[scale=1,width=0.8\textwidth]{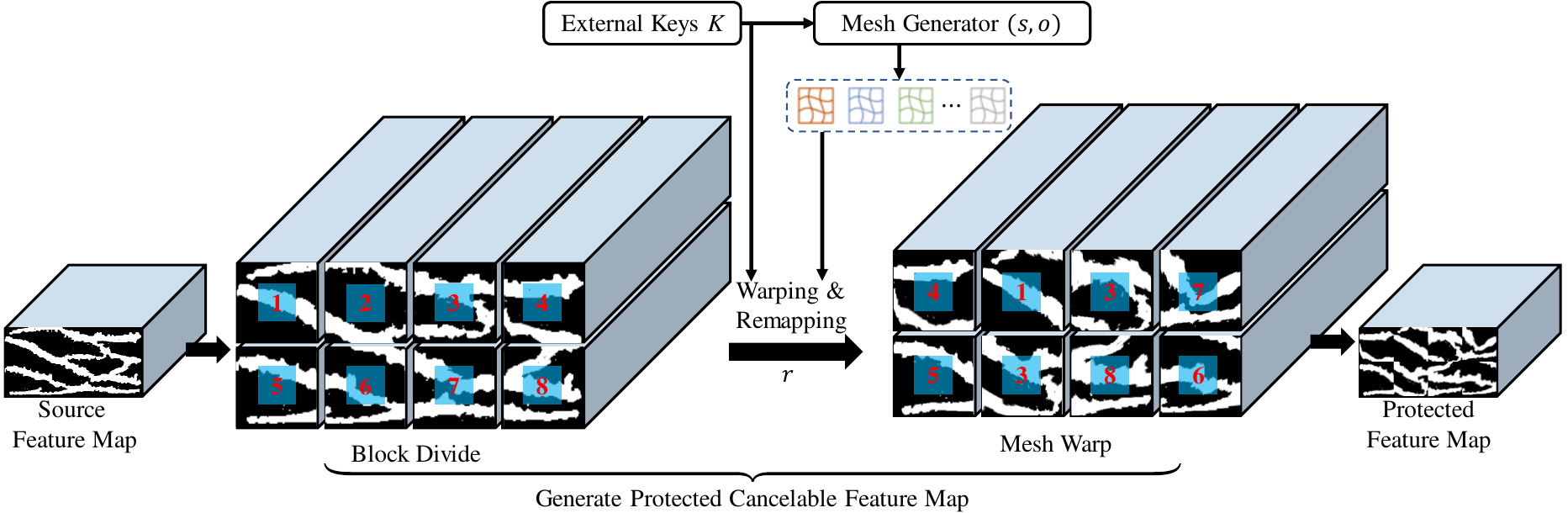}
		\caption{\centering{Transformation sub-module. $b$: block size, $s$: mesh size, $o$: warping factor and $r$: resampling rate (In this work, $b=16$, $s=8$, $o=0.625$, $r=0.8$).}}
		\label{fig6}
	\end{figure*}

	Inspired by group convolution \cite{9477103}, we divide the features into four
	groups and fuse each group, thereby reducing the redundant feature maps. Fig.
	\ref{fig5} shows the De-Redundant Convolution (De-R Conv), which is divided into
	two paths. First, for a feature map $F$ of size $(B,C,H,W)$, we use average
	pooling and two $1\times1$ Conv to obtain the weight $w$ and inverse
	weight
	$w^{\prime}$ of each channel (where $w^{\prime}$ is obtained by flipping $w$). This operation is similar to channel attention.
	Weights $w$ and $w^{\prime}$ are assigned to the corresponding feature maps to
	obtain $F_w$ and $F_{w^{\prime}}$ (upper and lower parts
	of Fig. \ref{fig6}), respectively, which are the positively and negatively weighted
	paths. Divide $F_w$ and $F_{w^{\prime}}$ into $4$ groups, and linearly fuse
	them to get the feature maps $F_w^\prime$ and $F_{w^{\prime}}^\prime$ of size
	$(B,\frac{C}{4},H,W)$ as
	
	\begin{equation}\label{eq5}	
		\left\{
		\begin{aligned}
			&F_w\Rightarrow\{F^1_w, F^2_w, F^3_w, F^4_w\},&\\
			&F_w^\prime=F^1_w\oplus F^2_w\oplus F^3_w\oplus F^4_w,&
		\end{aligned}
		\right.
	\end{equation}
	where $\Rightarrow$ denotes dividing the weighted feature maps equally into four
	groups of non-overlapping feature maps, $\oplus$ denotes that the four groups
	of feature maps are linearly fused at the corresponding positions. Next, we feed
	$F_w$ and $F_{w^{\prime}}$ into a $3\times3$ convolution to extract the
	dense feature maps. Note that 
	as $w^\prime$ is flipped
	for $F_{w^{\prime}}$
	in the reverse-weighted path, feature
	maps with lower responses are given higher weights,
	However, they contain rich features in the forward-weighted path. Hence,
	we use a $1\times1$ convolution to further compress them. Finally,
	the forward path has feature maps of size $(B,\frac{C}{4},H,W)$, and
	the reverse path has feature maps of size $(B,\frac{C}{8},H,W)$.
	The feature maps from both the forward and reverse paths are concatenated 
	(of size $(B,\frac{3C}{8},H,W)$)
	and used as outputs. Before inputting the features to the transformation module,
	rotation correction of the vein features is performed using the
	rotation correction parameter.
	\subsubsection{\textbf{Transformation}}
	
	In an unprotected FVR system, an attacker can invade the database 
	or analyze the network output of the finger vein information,
	In particular, for FVR systems based on DCNN, the main component of the
	DCNN is the convolution operation, and deconvolution can restore the operation
	to some extent and eventually restore the original feature template. This is the
	main security risk of DCNN-based FVR system. While we introduce feature
	transformation into the DCNN, the shallow features in the source domain
	are transformed to the protected domain by combining the user-defined external
	key, and the final identification of the user is performed in the protected
	domain. In BWR-ROIAlign, the transformation sub-module uses a
	block warping remapping (BWR) template protection method to realize the template
	transformation. This method is divided into two parts: block mesh warping and
	block remapping. It has four hyperparameters: image block size $
	b$, warping mesh size $s$, warping factor $o$, and resampling rate
	$r$. These parameters are system-level hyperparameters that are set prior to training the CFVNet, and are used to balance the
	recognition 
	system's
	template protection attributes 
	and recognition performance. The user's key corresponds to a unique tokenized pseudo-random feature transformation, which ensures that each user possesses a unique template protection transformation under the current parameter set, assuring accurate recognition.
	
	\begin{figure}[]
		\centering
		\captionsetup[subfloat]{labelsep=none,format=plain,labelformat=empty}
		\subfloat[   $G_T$]{
			\includegraphics[width=0.24\linewidth]{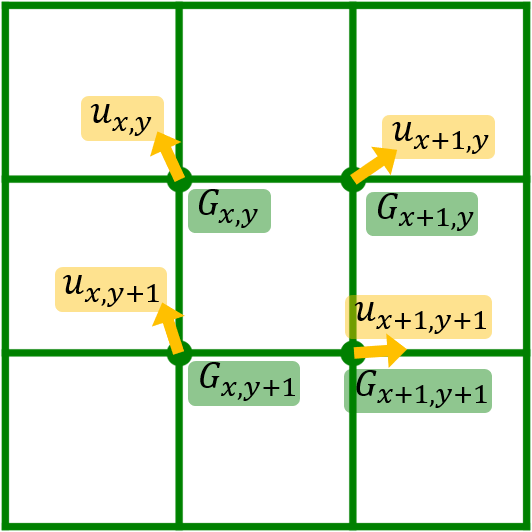}}
		\subfloat[]{
			\includegraphics[width=0.03\linewidth]{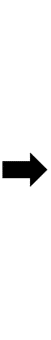}}
		\subfloat[   $B_T(A_T)$]{
			\includegraphics[width=0.24\linewidth]{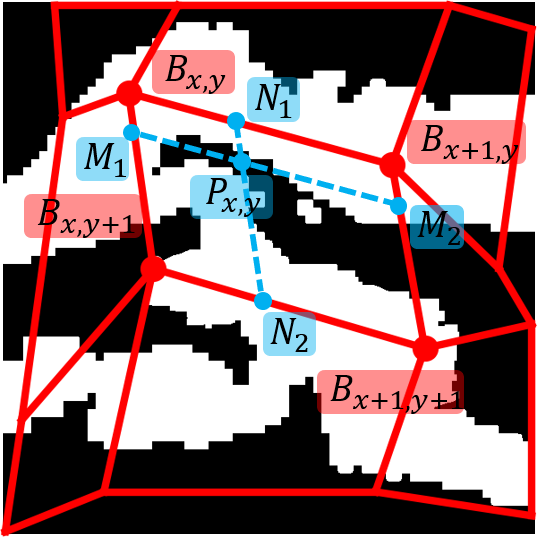}}
		\subfloat[]{
			\includegraphics[width=0.03\linewidth]{figure/Fig.8Arrow.png}}
		\subfloat[   $C_T$]{
			\includegraphics[width=0.24\linewidth]{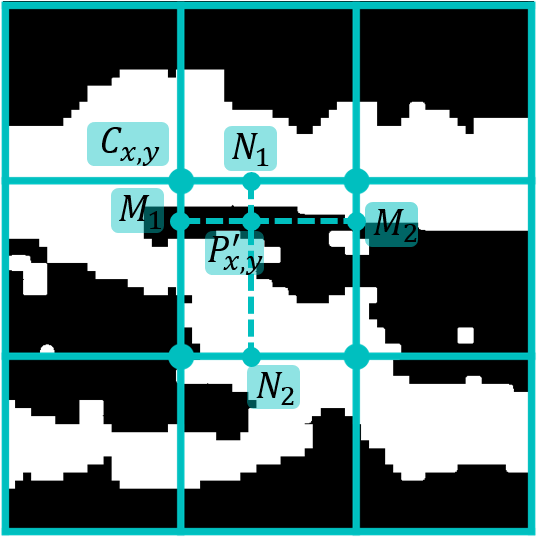}}
		\caption{\centering{Feature block warping. $G_T$ is the initialized mesh,
				$B_T(A_T)$ is the divided feature block with warped mesh, and $C_T$ is the
				warped feature block. $u_{x,y}$ is the transformation at vertex $G_{x,y}$,
				and warped vertex $B_{x,y} = G_{x,y} + u_{x,y}$, $u_{x,y}$ is a
				transformation in the vertex $G_{x,y}$ as in Eq. \ref{eq7}.
				$B_{x,y}B_{x+1,y}B_{x+1,y+1}B_{x,y+1}$ are the four vertices of a warped
				grid (abbreviated $B_1 B_2 B_3 B_4$ in the text). $M_1 M_2$ is the $t_x$th
				node line in the direction of $B_{x,y}B_{x+1,y}$, $N_1 N_2$ is the $t_y$th node line in the direction of $B_{x+1,y}B_{x+1,y+1}$. $P$ is the intersection of $M_1 M_2$ and $N_1 N_2$ in $B_T(A_T)$. $P^\prime$ is the point corresponding to $P$ in $C_T$.}}
		\label{fig7}
	\end{figure}
	
	Let the
	size 
	of the source domain feature map $F(x,y)$  be
	$(H,W)$. We divide $F$ into 
	$b\times b$
	pixel blocks:
	$\{A_T\}$, where $T=1,2,\dots,HW/b^2$. Each $A_T$ is assigned an
	initial $s\times s$ 
	quadrilateral mesh $G_T$. 
	The vertices of all 
	meshes in $G_T$ 
	are then randomly transformed 
	to obtain the warped mesh $B_T$. For any initial mesh, let
	$(G_{xi}, G_{yi})$ be the 
	coordinates of the $i$th
	vertex $G_i$.
	Similarly, let $B_i$ be the $i$th vertex after the transformation, and
	$(B_{xi}, B_{yi})$ be the 
	coordinates
	after transformation. The coordinates are related
	as
	\begin{equation}\label{eq6}	
		\left\{
		\begin{aligned}
			&B_{xi}=G_{xi}+\varDelta x\times o_x,&\\
			&B_{yi}=G_{yi}+\varDelta y\times o_y,&
		\end{aligned}
		\right.
	\end{equation}
	where $\varDelta x$ (resp. $\varDelta y$) is the length of single mesh in
	the $X$ (resp.
	$Y$) direction, $o_x$  (resp. $o_y$) is the warping factor in the $X$ (resp. $Y$)
	direction (and takes values in $[0,o]$). 
	The 
	higher the warping degree, 
	the closer is its value to $1$,
	and vice versa.
	
	After the vertices are transformed, all $A_T$'s are divided by randomly warped
	irregular meshes. These meshes are then mapped to the corresponding regular
	meshes of the new feature block $C_T$'s. Consider a given warped grid $B_1 B_2 B_3
	B_4$ (with edges $B_1 B_2 ,$ $B_2 B_3 ,$ $B_3 B_4$ and $B_4 B_1$). These
	edges are divided into $t-1$ equal parts. $B_1 B_2 B_3
	B_4$ is divided into $t \times t$
	nodes. Similarly, a grid of the new feature block is also divided into $t \times t$ nodes.
	Let $M_1 M_2$ be the $t_x$th node line in the direction of $B_1 B_2$, 
	$N_1 N_2$ be the $t_y$th node line in the direction of $B_2 B_3$ (where $t_x,t_y
	\in [0,t]$). Let $P$ be the intersection of $M_1 M_2$ and $N_1 N_2$. The 
	coordinates $(P_x, P_y)$ of $P$ are
	\begin{equation}\label{eq7}	
		\left\{
		\begin{aligned}
			&P_x=M_{1x}+\frac{M_{2x}-M_{1x}}{t}\times t_x,&\\
			&P_y=M_{1y}+\frac{M_{2y}-M_{1y}}{t}\times t_x,&
		\end{aligned}
		\right.
	\end{equation}
	where $(M_{1x}, M_{1y})$ and $(M_{2x}, M_{2y})$ are the 
	coordinates of $M_1$ and $M_2$, respectively. Next, the point $P$ is mapped to the same position in the grid of the new feature block as point $P^\prime$. The 
	coordinates $(P_x^\prime, P_y^\prime)$ of $P^\prime$ are
	\begin{equation}\label{eq}	
		\left\{
		\begin{aligned}
			&P_x^\prime=C_x+\varDelta t\times t_x,&\\
			&P_y^\prime=C_y+\varDelta t\times t_y,&
		\end{aligned}
		\right.
	\end{equation}
	where $C$ is the top-left vertex of the grid of the new feature block, $(C_x,
	C_y)$ are the 
	coordinates of $C$, and $\varDelta t$
	is the length between the two points in the new feature block after
	subdivision. Fig. \ref{fig7} shows the feature block warping process.
	The $C_T$'s 
	are sampled 
	with a sampling rate $r$. 
	A subset $\{C_T^\prime\}$ of $\{C_T\}$ 
	is selected 
	to reconstruct the image, and 
	$\{C_T^\prime\}$ 
	is randomly mapped 
	to
	the feature map $F_E$ to produce the final protected domain feature map (note that
	some blocks in $\{C_T^\prime\}$ will be reused) in the following way:
	\begin{equation}\label{eq9}	
		F_E=F_{map}(f_r(\{C_T\},r),k),
	\end{equation}
	where
	$f_r(\cdot)$ denotes random sampling the warped image block, $k$ is a user's key, and $f_{map}(\cdot)$ denotes random sampling of the subset
	$\{C_T^\prime\}$ obtained after the operation of $f_r(\cdot)$, which is then mapped to
	generate the protected feature map $F_E$. 
	
	The transformation sub-module 
	introduces unlinkability and irreversibility to the recognition
	system, which ensures that the subsequent feature extraction and recognition are  
	performed in the protected domain. It should
	be noted that when training the CFVNet model, all users have
	unique and correct keys. When using the trained model for
	inference/recognition, enrolled users with the correct finger vein features and
	keys will get high confidence in recognition. Using non-enrolled user's finger
	vein features or incorrect key is considered an illegal attack, and the
	model will output very low confidence. Once the user's key is stolen, 
	it is necessary to assign a new key to the user and retrain the CFVNet model
	for security.
	
	\section{Experiments}
	\label{sec:expt}
	
	In this section, the proposed CFVNet is used for FVR. We perform a number of
	experiments on multiple datasets.
	Section~\ref{sec:setup}
	first describes the experimental setup, including the finger vein
	datasets, evaluation metrics and implementation details. 
	In Section~\ref{sec:4-3},
	ablation
	experiments are designed to validate effectiveness of BWR-ROIAlign.
	Next, experiments are conducted to analyze the
	effectiveness of the three sub-modules in BWR-ROIAlign
	on localization 
	(Section~\ref{sec:local}),
	model size and FLOPS (Section~\ref{sec:size}), unlinkability, revocability
	and irreversibility (Section~\ref{sec:4-6}). Finally, 
	in Section~\ref{sec:sota}, we
	compare with various FVR works and  shows
	the advantages of CFVNet in recognition and template protection.
	
	\subsection{Experimental Setup} \label{sec:setup}
	\subsubsection{Datasets}
	Four datasets are used (i) SDUMLA-FV
	\cite{yin2011sdumla}, a multimodal feature dataset from Shandong University, (ii)
	the UTFVP \cite{ton2013high} dataset from the University of Twente, (iii) the FV-USM
	\cite{asaari2014fusion} dataset from the Universiti Teknologi Malaysia, and (iv)
	the HKPU-FV \cite{kumar2011human} dataset from the Hong Kong Polytechnic University.
	Table \ref{tb1} shows the information of each dataset. Note that the
	finger vein image sizes in all datasets are normalized to $256 \times 320$ (the
	images in FV-USM are first flipped to finger horizontally, and then centrally cropped to $360 \times 480$).
	
	\begin{table}[]
		\caption{Details of the experimental datasets.}
		\renewcommand\arraystretch{0.2}
		\label{tb1}
		\centering
		\setlength\tabcolsep{3pt} 
		\begin{tabular}{cccccc}
			\toprule
			Datasets  & Subjects & Fingers & \begin{tabular}[c]{@{}c@{}}Capture\\ times\end{tabular} & \begin{tabular}[c]{@{}c@{}}Total \\ images\end{tabular} & \begin{tabular}[c]{@{}c@{}}Image\\ size\end{tabular} \\ \midrule
			HKPU-FV   & 250      & 6       & 12/6                                                    & 3132                                                    & 513×256pxl                                           \\
			SDUMLA-FV & 106      & 6       & 6                                                       & 3816                                                    & 320×240pxl                                           \\
			UTFVP     & 60       & 6       & 4                                                       & 1440                                                    & 672×380pxl                                           \\
			FV-USM    & 123      & 4       & 6                                                       & 5904                                                    & 640×480pxl                                           \\ \bottomrule
		\end{tabular}
	\end{table}
	
	\subsubsection{Evaluation Metrics}
	To evaluate FVR performance, commonly used metrics include (i) false accept
	rate (FAR), 
	which is the probability that the recognition system accepts a fake as enrolled user;
	(ii) false reject rate (FRR), 
	which is the probability that the system rejects an enrolled user as a fake; 
	(iii) equal error rate (EER),
	which is the point where FAR equals to FRR;
	and (iv)
	receiver operating characteristic (ROC) curve. 
	
	
	
	The localization performance is evaluated by intersection over
	union (IoU) \cite{svanstrom2021real}, which is commonly used to measure the overlap between the predicted box and its true annotation:
	\begin{equation}\label{eq10}	
		IoU=\frac{Area~ of~ Overlap}{Area~ of~ Union},
	\end{equation}
	where $Area\ of\ Overlap$ and $Area\ of\ Union$ denote the intersection area and concatenation area of the prediction box and its true annotation, respectively.
	
	Unlinkability of a cancelable recognition system is assessed based on the
	following global unlinkability metric 
	\cite{gomez2017general}:
	\begin{equation}\label{eq11}	
		D_\leftrightarrow^{sys}=\int p(s|H_m)[p(H_m|s)-p(H_{nm}|s)]ds,
	\end{equation}
	where 
	$s$ is the linkage score between 
	two protected templates $T_1$ and $T_2$.
	Based on the paired hypothesis
	$H_m=$``both templates belong to mated
	instances" and the unpaired hypothesis $H_{nm}=$``both templates belong
	to\ non-mated instances", we obtain 
	the conditional probabilities
	$p(H_m|s)$ and $p(H_{nm}|s)$, 
	for a given matching score $s$. $p(s|H_m)$ is the conditional probability of
	obtaining a score $s$ knowing that the two templates come from mated instances. A small $D_\leftrightarrow^{sys}$ indicates that the link is weak, and vice versa.
	
	Invertibility 
	is evaluated by
	the pre-image attack match rate (P-IAMR) \cite{gomez2020reversing}, which is
	defined as the acceptance rate of a protected system when accessed with the original biometric feature samples from a subject:
	\begin{equation}\label{eq12}	
		\text{P-IAMR}=\frac{1}{N}\sum\nolimits_{i=1}^{N}\delta\{y_i=y_{gt}\},
	\end{equation}
	where $N$ is the number of pre-image attacks, $y_i$ is the predicted class label
	of the current attack sample, and $y_{gt}$ is its corresponding true class label.
	The function $\delta\{\text{condition}\}$ is $1$ if the condition is satisfied, and $0$ otherwise.
	A higher P-IAMR indicates a higher level of invertibility for the template. 
	
	The recognition system is revocable if the impostor and pseudo-impostor
	distributions overlap and the genuine and pseudo-impostor distributions are
	clearly distinguishable. The degree of overlap between two distributions
	can be measured using the \textit{decidability index} $d^\prime$, which
	reflects the revocability of the recognition system
	\cite{sadhya2019generation}. This metric is described as the normalized
	distance between the means of the two distributions. 
	For two distributions with means $\mu_1$ and $\mu_2$, and standard deviations $\sigma_1$ and $\sigma_2$, respectively, $d^\prime$ is defined as
	\begin{equation}\label{eq13}	
		d^\prime = \frac{|\mu_1 - \mu_2|}{\sqrt{\frac{1}{2}(\sigma_1^2 +
				\sigma_2^2)}}.
	\end{equation}
	A higher value of $d^\prime$ indicates good separation between the two distributions, and vice versa.
	
	\subsubsection{Implementation details}
	
	To train the CFVNet, we adopt Adam \cite{kingma2014adam} as the optimizer (with a momentum of
	0.9). The loss consists of the classification loss (cross-entropy loss) and regression
	loss (smooth $\ell_1$ loss). The learning rate and batch size are set to 0.001 and
	64, respectively. The CFVNet model is trained on two RTX 3090 GPUs, Intel Xeon(R)
	sliver 4110 CPU@2.10GHz and 32GB RAM using the PyTorch framework for 200 training
	epochs. All datasets are divided into training and testing sets in the ratio of
	$8:2$.
	
	\subsection{Ablation Experiments}
	\label{sec:4-3}

	\begin{figure*}[t]
		\centering
		\hspace{-5mm}
		\subfloat[\label{fig:a2}]{
			\includegraphics[width=0.24\linewidth]{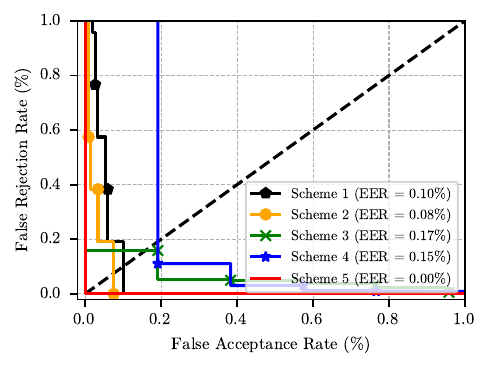}}
		\hspace{-3mm}
		\subfloat[\label{fig:b2}]{
			\includegraphics[width=0.24\linewidth]{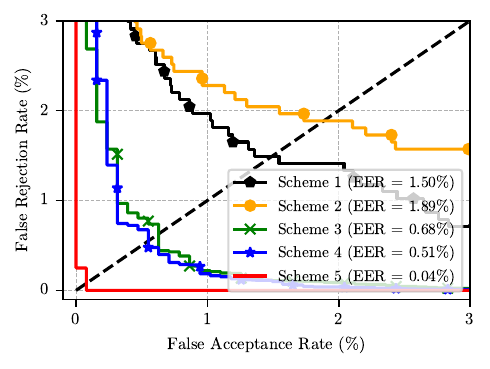}}
		\hspace{-2mm}
		\subfloat[\label{fig:c2}]{
			\includegraphics[width=0.24\linewidth]{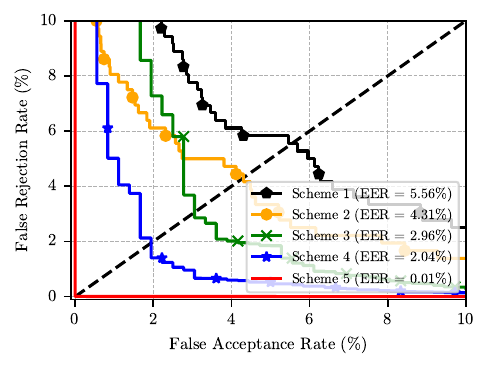}}
		\hspace{-2mm}
		\subfloat[\label{fig:d2}]{
			\includegraphics[width=0.24\linewidth]{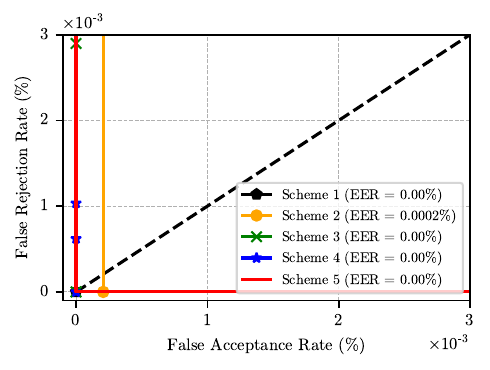}}
		\caption{\centering{Comparison of ablation experiment ROC curves for five schemes on each dataset. (a) HKPU-FV (b) SDUMLA-FV (c) UTFVP (d) FV-USM}}
		\label{fig12} 
	\end{figure*}
	
	In this section, we perform ablation experiments on CFVNet. Five schemes are
	included 1) Directly using the raw images (size $256 \times 320$) to
	train a standard DCNN; 2) Using manually ROI images (size $128 \times 256$) to
	train a standard DCNN; 3) Perform spatial redundancy removal by activating only
	the localization sub-module in BWR-ROIAlign; 4) Fully activating the compression
	sub-module on top of scheme 3; 5) Activate the localization, compression and transformation sub-modules in BWR-ROIAlign module.

	\begin{table}[h]
		\caption{Comparison among ablation experiments recognition performance (\%) of the five schemes on each dataset.}
		\label{tb3}
		\centering
		\renewcommand\arraystretch{0.8}
		\setlength\tabcolsep{2pt}
		\begin{tabular}{ccccccccccc}
			\toprule
			\multirow{2}{*}{Scheme} & \multicolumn{2}{c}{HKPU-FV}                         & \multicolumn{2}{c}{SDUMLA-FV}                       & \multicolumn{2}{c}{UTFVP}                           & \multicolumn{2}{c}{FV-USM}       & \multicolumn{2}{c}{Average}     \\ \cmidrule{2-11} 
			& \multicolumn{1}{c|}{ACC} & \multicolumn{1}{c|}{EER} & \multicolumn{1}{c|}{ACC} & \multicolumn{1}{c|}{EER} & \multicolumn{1}{c|}{ACC} & \multicolumn{1}{c|}{EER} & ACC   & \multicolumn{1}{c|}{EER} & \multicolumn{1}{c|}{ACC} & EER  \\ \midrule
			1                 & 97.89                    & 0.10                     & 91.90                    & 1.50                     & 61.94                    & 5.56                     & \textbf{100}   & \textbf{0.00}                     & 87.93                    & 1.79 \\
			2                 & 98.66                    & 0.08                     & 91.35                    & 1.89                     & 75.28                    & 4.31                     & 99.89 & 0.0002                   & 91.30                    & 1.57 \\
			3                 & 98.66                    & 0.17                     & 95.79                    & 0.68                     & 80.56                    & 2.96                     & \textbf{100}   & \textbf{0.00}                   & 93.75                    & 0.95 \\
			4                 & 98.66                    & 0.15                     & 96.46                    & 0.51                     & 84.72                    & 2.04                     & \textbf{100} & \textbf{0.00}                     & 94.96                    & 0.68 \\
			5                 & \textbf{100}                      & \textbf{0.00}                     & \textbf{99.84}                    & \textbf{0.04}                     & \textbf{99.44}                    & \textbf{0.01}                     & \textbf{100}   & \textbf{0.00}                     & \textbf{99.82}                    & \textbf{0.01} \\ \bottomrule
		\end{tabular}
	\end{table}	
	
	\begin{table}[t]
		\renewcommand\arraystretch{0.5}
		\setlength\tabcolsep{2.1pt}
		\centering
		\begin{threeparttable}
			\caption{Comparison among rotation correction effects.}
			\label{tb4}
			\begin{tabular}{ccccccccc}
				\toprule
				\multirow{2}{*}{} & \multicolumn{2}{c}{HKPU-FV}                         & \multicolumn{2}{c}{SDUMLA-FV}                       & \multicolumn{2}{c}{UTFVP}                           & \multicolumn{2}{c}{FV-USM}        \\ \cmidrule{2-9} 
				& \multicolumn{1}{c|}{ACC} & \multicolumn{1}{c|}{EER} & \multicolumn{1}{c|}{ACC} & \multicolumn{1}{c|}{EER} & \multicolumn{1}{c|}{ACC} & \multicolumn{1}{c|}{EER} & \multicolumn{1}{c|}{ACC} & EER    \\ \midrule
				CFVNet-L\tnote{*}         & 98.85                    & 0.38                     & 94.97                    & 0.79                     & 78.33                    & 4.17                     & 100                      & 0.0002 \\
				CFVNet-L          & 98.66                    & 0.17                     & 95.79                    & 0.68                     & 80.56                    & 2.96                     & 100                      & 0.0    \\ \midrule
				CFVNet-S\tnote{*}         & 98.66                    & 0.19                     & 95.13                    & 0.55                     & 77.78                    & 2.78                     & 99.80                    & 0.0    \\
				CFVNet-S          & 98.66                    & 0.15                     & 96.46                    & 0.51                     & 84.72                    & 2.04                     & 100                      & 0.0    \\ \midrule
				CFVNet\tnote{*}           & 100                      & 0.0                      & 99.37                    & 0.06                     & 98.61                    & 0.02                     & 100                      & 0.0    \\
				CFVNet            & 100                      & 0.0                      & 99.84                    & 0.04                   & 99.44                    & 0.01                     & 100                      & 0.0    \\ \bottomrule

			\end{tabular}
			\begin{tablenotes}
				\item[*] No regression on $\varphi$ and no rotation alignment was performed.
			\end{tablenotes}
		\end{threeparttable}
	\end{table}
	
	Fig. \ref{fig12} shows the ROC curves 
	obtained on each dataset. It can be observed that CFVNet's ROC
	curve is very close to the coordinate axis, and achieves very low EERs compared
	to the other schemes. Table \ref{tb3} shows the recognition accuracy and EER
	obtained on each dataset. 
	As can be seen, 
	using manually extracted ROI images to train the standard DCNN
	improves
	the average recognition accuracy 
	by 3.37\% and 
	reduces 
	the average EER  by
	0.22\% compared to
	Scheme 1. This is due to the fact that the feature information used for analysis
	and discrimination in the extracted ROI finger vein image is more stable, and the
	complex background noise of finger shape, position or device is removed. These 
	reduce the difference between the same instances to some extent. In Scheme 3,
	we activate the localization sub-module in BWR-ROIAlign and activate spatial
	redundancy removal in the compression sub-module. The average recognition
	accuracy is improved by 2.45\% and the average EER is reduced by 0.62\%
	compared with
	Scheme 2. Unlike manual ROI extraction which achieves hard alignment of the stable
	regions, the localization sub-module can be seen as soft feature alignment.
	Benefiting from the correction of the recognition branch, the
	localization branch's parameters are optimized towards the smallest differences in
	the same instances (which also explains why the proposed model is not the most compatible
	approach with manually labelled regions among other ROI-localizable DCNN models).
	Scheme 4 improves the average recognition accuracy 
	by 1.21\% and 
	reduces 
	the average EER  by
	0.27\% while reducing nearly 4/5 of the parameters. In Scheme 2, although spatial redundancy
	is removed to make the analysis discriminant region more stable, there are still unstable tissue features around the vein, and the redundant feature
	channels may overly focus on changes in these regions. Using De-R Conv in the
	compression sub-module to remove the redundant channels allows the model to focus
	more on the vein pattern information. In Scheme 5, the transformation
	sub-module completes the transformation from source-domain features to
	protected-domain features, and then extract and analyze the protected-domain
	features, finally completing the recognition process in protected-domain by using the
	transformed features. Usually when a biometric template is protected, the
	biometric information contained in the template (such as vein branches,
	intersections, and branching angles) is destroyed due to the transformation
	operations. The recognition system performance, which use protected templates,
	decreases dramatically. However, comparing with the baseline model, the average
	recognition accuracy of CFVNet is improved by 11.89\% and the average EER is reduced
	by 1.78\%. This is due to the powerful learning ability of DCNN. Traditional
	recognition systems based on manual processing of features is unable
	to deal with this abstract transformation, while the deeper convolutional layer in
	CFVNet learns the feature mapping rules in the transformation sub-module. This
	allows
	CFVNet to have significantly improved performance.

	\subsection{Localization Performance Analysis}
	\label{sec:local}
		\begin{figure}[]
		\centering
		\captionsetup[subfloat]{labelsep=none,format=plain,labelformat=empty}
		\subfloat[ Standard DCNN]{
			\includegraphics[width=0.3\linewidth]{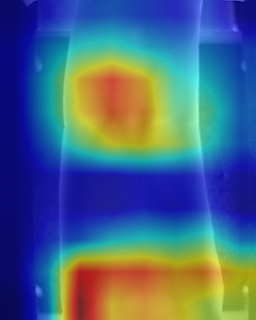}}\label{fig:a3}
		\subfloat[ CFVNet]{
			\includegraphics[width=0.3\linewidth]{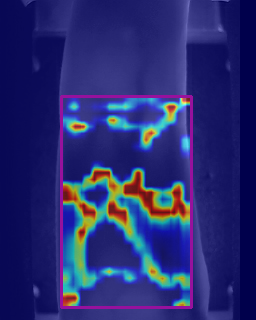}}\label{fig:b3}
		\caption{\centering{Standard DCNN and CFVNet-based class activation maps.}}
		\label{fig13} 
	\end{figure}
	In this section, we study the CFVNet-based recognition system using class activation
	maps (CAM) \cite{selvaraju2017grad} and localization performance. Fig.
	\ref{fig13} compares the CAMs obtained on training a standard DCNN and CFVNet using raw images. 
	When the raw image is used for recognition
	(Fig. \ref{fig13}, left), 
	the model focuses on the feature information in the whole image (including
	device background, finger position or shape, surrounding tissue region and vein
	pattern). However, the feature region used for discrimination is fuzzy and unstable, and 
	even 
	non-vein region features 
	are utilized 
	for classification. 
	After using the BWR-ROIAlign module through localization and compression
	(Fig. \ref{fig13}, right), 
	unstable noise due to the device and acquisition environment is removed,
	and compressing the redundant channels allows the model to focus more on the vein
	pattern information rather than the other surrounding tissues.
	
	Table \ref{tb5} shows the localization effect. 
	CFVNet-L, CFVNet-S and CFVNet correspond to Schemes 3-5 in
	Section~\ref{sec:4-3}, respectively. As can be
	seen, the average IoUs are 89.53, 89.02 and 89.09, respectively. Common object
	detection models can obtain high IoUs due to their excellent object searching
	abilities. However, classification confidence tends to be low because these
	models mainly look for objects that may exist in the image. This is not
	suitable for the task of identification (due to low classification confidence,
	which leads to high FAR and FRR). In CFVNet, due to the
	penalty of the recognition branch to feature alignment in the localization branch,
	the ROI is aligned towards the smallest differences in the same instances instead
	of manually labeling the regions. This further proves that the localization
	sub-module is a soft feature alignment scheme. Similarly, this soft alignment is
	compatible with the more refined feature alignment required by the feature
	transformation during the source to protected domain.
	
	\begin{table}[]
		\caption{\centering{Comparison localization performance on each dataset.}}
		\label{tb5}
		\centering
		\setlength\tabcolsep{3pt}
		\renewcommand\arraystretch{0.5}
		\begin{tabular}{ccccc}
			\toprule
			\multirow{2}{*}{Model} & \multicolumn{4}{c}{IoU}             \\ \cmidrule{2-5} 
			& HKPU-FV & SDUMLA-FV & UTFVP & FV-USM \\ \midrule
			Faster RCNN            & 91.66   & 92.34    & 85.12 & 91.74  \\
			SSD300                 & 90.74   & 93.12    & 91.24 & 94.19  \\
			CFVNet-L                 & 86.36   & 90.74    & 87.72 & 93.31  \\
			CFVNet-S                 & 86.83   & 90.06    & 85.95 & 93.23  \\
			CFVNet                   & 85.45   & 89.73    & 86.33 & 84.85  \\ \bottomrule
		\end{tabular}
	\end{table}

	\subsection{Model Size and Runtime Analysis}
	\label{sec:size}
		\begin{table}[]
		\caption{Comparison among GFLOPs, Params and Inference Time.}
		\label{tb6}
		\centering
		\renewcommand\arraystretch{0.5}
		\setlength\tabcolsep{2.9pt}
		\begin{tabular}{cccccc}
			\toprule
			Model              & GFLOPs & Params (M) & Time (ms) & ACC (\%) & EER (\%) \\ \midrule
			MobileNet          & 1.48   & \textbf{4.97}       & 16.33     & 56.92    & 5.69     \\
			ResNet50           & 6.62   & 24.73      & 16.84     & 87.93    & 1.79     \\
			EfficientNet-B7    & 34.66  & 65.32      & 52.52     & 96.73    & 0.35     \\
			VIT & 39.86  & 203.36     & 15.92     & 69.42    & 5.88     \\
			SSD300             & 34.86  & 35.64      & 103.86    & 89.42    & 48.17    \\
			Faster RCNN        & 167.98 & 41.76      & 34.56     & 92.32    & 43.73    \\ \midrule
			CFVNet-L             & 2.70   & 38.58      & \textbf{14.39}     & 93.75    & 0.95     \\
			CFVNet-S             & \textbf{0.35}   & 8.46       & 15.24     & 94.96    & 0.68     \\
			CFVNet               & \textbf{0.35}   & 8.46       & 31.26     & \textbf{99.82}    & \textbf{0.01}     \\ \bottomrule
		\end{tabular}
	\end{table}
	Table \ref{tb6} compares mainstream classification models and CFVNet models in terms
	of GFLOPs, number of parameters (Params), inference time and average
	recognition performance (the number of
	classes  is 600). Except MobileNet model that cannot achieve 
	effective recognition task, the proposed CFVNet has the smallest number of model
	parameters and the lowest computational consumption. Moreover, the recognition
	performance is significantly improved compared with the other models. CFVNet-L
	takes
	the least inference time compared with the other models, and CFVNet-S reduces 
	the parameters 
	by $4/5$
	by non-destructively removing  the redundant feature
	channels. 
	The inference time is increased only by 0.74 ms. In CFVNet, the transformation of
	the vein feature source domain to the protected domain is achieved by utilizing a
	feature cancelable transformation proposed in the transformation sub-module. In template protection tasks, this operation
	usually requires more time. The feature maps obtained in DCNN are
	usually high-dimensional, which further increases the burden of the transformation
	sub-module. Hence, we remove the redundant feature channels by the compression
	sub-module, so as to keep CFVNet's inference time to be relatively small and
	still
	improve recognition efficiency.
	
	\subsection{Unlinkability, Revocability and Irreversibility Analysis}
	\label{sec:4-6}
	In this section, we evaluate the template protection attributes and security of the CFVNet-based FVR system on four datasets and their mixed datasets. We consider both the normal scenario and the stolen scenario. In the normal scenario, which is the expected scenario for most cases, the hyperparameter settings and the user's key are considered secret and not disclosed. However, in the stolen scenario, the attacker (impostor) has the system's hyperparameters 
	and/or the enrolled user's key to access the system in combining with the
	unenrolled finger vein template. 
	The following three scenarios are
	considered. 1) Normal scenario, where the attacker does not have any
	information about the recognition system and accesses the system using an
	unenrolled template. 2) Transformation sub-module hyperparameters are stolen,
	where the attacker has the transformation sub-module hyperparameters of the target
	recognition system, and uses the hyperparameters to randomly generate a key to
	access the system combining with an unenrolled template. 3) The enrolled user's
	key is stolen, where the attacker has the enrolled user's right key and accesses
	the system combining with an unenrolled template. These scenarios pose gradually
	increasing security risks from 1 to 3. Based on these three scenarios, the template protection attributes and security of the CFVNet-based finger vein recognition system are further analyzed.
	\begin{table}[]
		\centering
		\renewcommand\arraystretch{0.5}
		\setlength\tabcolsep{3pt}
		\caption{Global linkability ($D_\leftrightarrow^{sys}$) in three scenarios on each dataset.}
		\label{tb7}
		\begin{tabular}{cccccc}
			\toprule
			Scheme & HKPU-FV & SDUMLA-FV & UTFVP & FV-USM & MIX-D \\ \midrule
			1      & 0.028   & 0.026     & 0.022 & 0.025  & 0.014 \\
			2      & 0.162   & 0.155     & 0.168 & 0.061  & 0.199 \\
			3      & 0.238   & 0.179     & 0.228 & 0.129  & 0.250 \\ \bottomrule
		\end{tabular}
	\end{table}

	Protected templates generated using different keys are unlinkable, which makes it
	possible to set up different keys to produce unlinkable protected templates when
	the user is enrolled in multiple applications. So we set five random keys for each instance to cross matching, record the mated scores and
	non-mated scores, and compute the global linkability $D_\leftrightarrow^{sys}$. It should be noted that during the evaluation process, in the normal scenario these keys are random keys generated under random hyperparameters, in scenario 2 the hyperparameters are stolen, which means that these keys are random keys generated under the hyperparameters of the target recognition system, and in scenario 3 the keys have been stolen and these keys are the correct keys of the enrolled users.

	\begin{figure}[t]
		\centering
		\includegraphics[scale=1,width=0.4\textwidth]{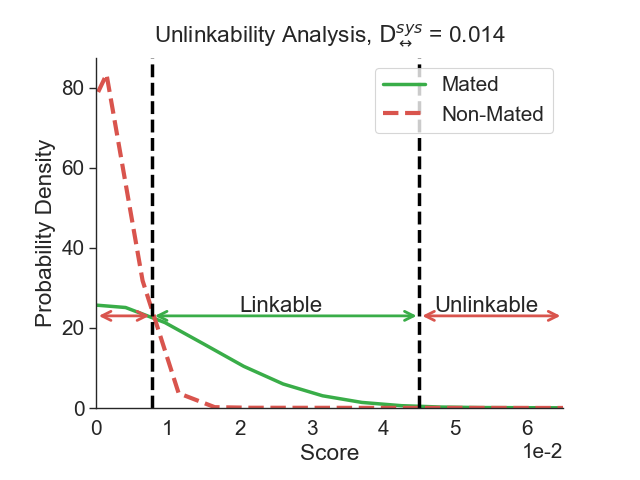}
		\caption{\centering{Normal scenario unlinkability analysis on MIX-D.}}
		\label{fig14}
	\end{figure}
	
	Table \ref{tb7} shows the global linkability metric $D_\leftrightarrow^{sys}$ on
	each dataset. In the normal scenario, the CFVNet-based finger vein recognition system
	presents minimal linkability and satisfies the CB requirement for unlinkability.
	In Scenario 2, when the system hyperparameters are stolen, linkability between the
	cross-systems slightly increases. In Scenario 3, the attacker has the enrolled
	user‘s key,  linkability between the cross-systems increases and shows some
	linkability. As can be seen from Fig. \ref{fig14}, there is considerable
	overlap between the mated and non-mated instance score distributions, which
	means that they have essentially the same distribution and are impossible to
	distinguish. Only when the mated scores are larger than the non-mated scores,
	the recognition system has minimal linkage shown in less than $3.7\%$
	area.

	\begin{figure}[t]
		\centering
		\includegraphics[scale=1,width=0.4\textwidth]{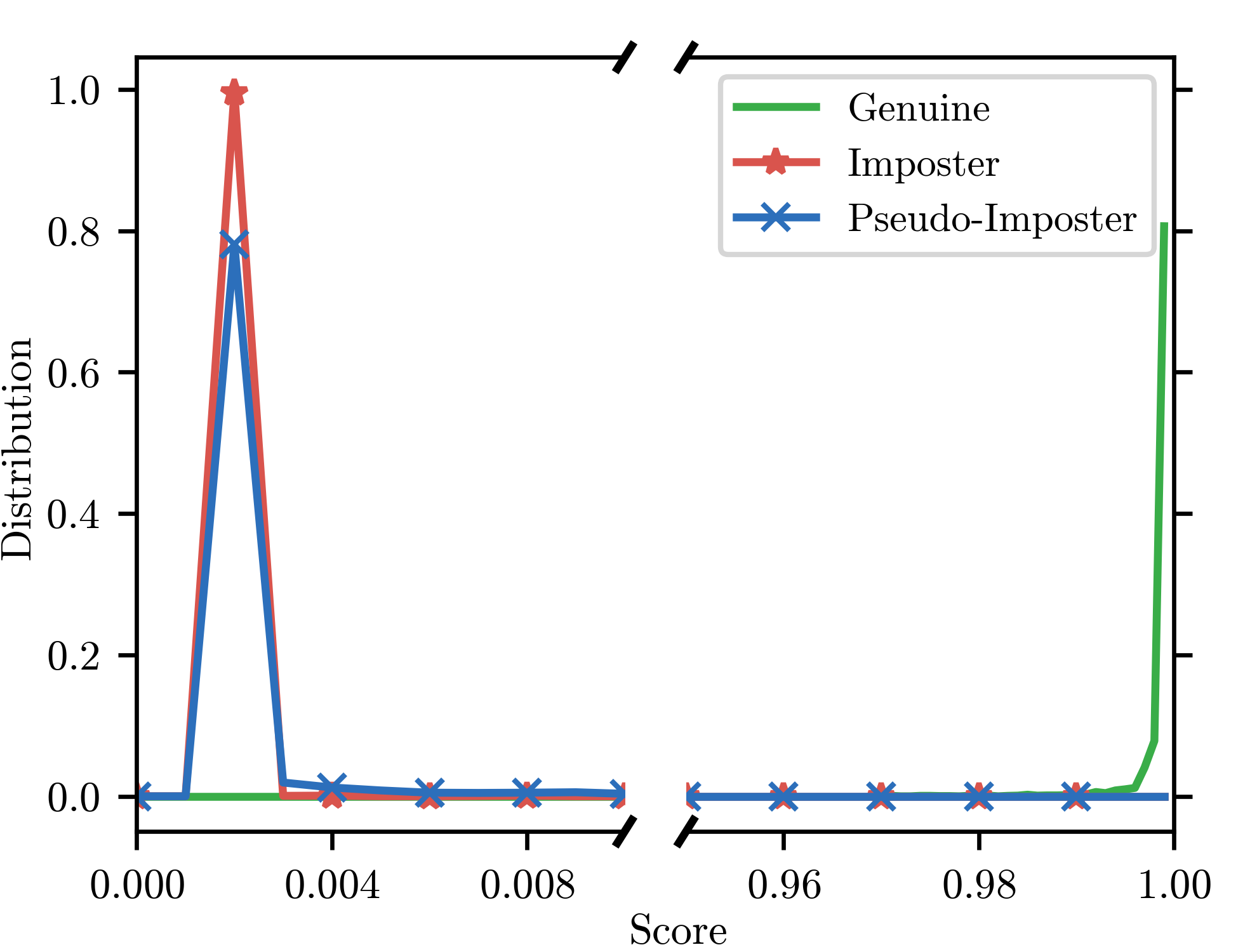}
		\caption{\centering{In normal scenario, The Genuine, Imposter and Pseudo-Imposter distribution on MIX-D.}}
		\label{fig15}
	\end{figure}
	
	Once a biometric is compromised, security of the recognition system decreases
	drastically. Cancelable biometrics ensure that templates generated using
	different keys are completely new. In other words, in CFVNet, the original (stolen)
	and new (regenerated) templates are generated from the same instances but the two
	templates are not linked. Revocability can empirically investigate this
	guarantee by generating the pseudo-impostor distribution. Similar to the
	evaluation of unlinkability, we set five random keys for each instance to cross matching. Cross-system impostor distributions are
	denoted as pseudo-impostor distributions. Revocability can be
	empirically validated if (1) the impostor and pseudo-impostor distributions
	overlap, and (2) the genuine and pseudo-impostor distributions have clear
	separability \cite{sadhya2019generation}. The genuine, impostor and
	pseudo-impostor scores distributions for MIX-D are shown in Fig. \ref{fig15}.
	There is a large overlap between the impostor and pseudo-impostor score
	distributions, whereas the genuine and pseudo-impostor scores are clearly
	distinguished. The separability or overlap between two distributions can be
	quantitatively estimated by the \textit{decidability index}
	$d^\prime$.
	Table \ref{tb8} shows the $d^\prime$ values between the three distributions on
	each dataset for different scenarios. In the normal scenario, the genuine score
	distribution is clearly distinguishable from the impostor and pseudo-impostor
	scores, while the impostor and pseudo-impostor score distributions highly
	overlap, which means that different protected templates generated from the same
	instances are similarly characterized to the inter-class template samples. The CFVNet-based recognition system satisfies revocability. In Scenarios 2 and 3, revocability is slightly compromised because the attacker has some information about the recognition system, but the revocability requirement is still satisfied. Therefore, when the recognition system is attacked or damaged, a new recognition system can be trained by re-issuing new hyperparameters and keys.

	\begin{table}[]
		\centering
		\caption{Decidability index ($d^\prime$) between the three distributions for the three scenarios on each dataset. $d^\prime_{GI}$, $d^\prime_{GP}$ and $d^\prime_{IP}$ are the discriminability indices between genuine/impostor score distributions, genuine/pseudo-impostor score distributions and impostor/pseudo-impostor score distributions.}
		\label{tb8}
		\renewcommand\arraystretch{0.1}
		\setlength\tabcolsep{2.9pt}
		\begin{tabular}{ccccc}
			\toprule
			\multicolumn{2}{c}{Scheme}     & 1     & 2     & 3     \\ \midrule
			\multirow{3}{*}{HKPU-FV}   & $d^\prime_{GI}$ & 501   & 501   & 501   \\ \cmidrule{2-2}
			& $d^\prime_{GP}$ & 2.84  & 1.87  & 1.38  \\ \cmidrule{2-2}
			& $d^\prime_{IP}$ & 0.36  & 0.84  & 1.19  \\ \midrule
			\multirow{3}{*}{SDUMLA-FV} & $d^\prime_{GI}$ & 26.42 & 26.42 & 26.42 \\ \cmidrule{2-2}
			& $d^\prime_{GP}$ & 9.00  & 1.65  & 1.46  \\ \cmidrule{2-2}
			& $d^\prime_{IP}$ & 0.27  & 0.89  & 1.04  \\ \midrule
			\multirow{3}{*}{UTFVP}     & $d^\prime_{GI}$ & 10.47  & 10.47  & 10.47  \\ \cmidrule{2-2}
			& $d^\prime_{GP}$ & 3.27  & 2.04  & 1.85  \\ \cmidrule{2-2}
			& $d^\prime_{IP}$ & 0.34  & 0.75  & 0.89  \\ \midrule
			\multirow{3}{*}{FV-USM}    & $d^\prime_{GI}$ & 57.28  & 57.28  & 57.28  \\ \cmidrule{2-2}
			& $d^\prime_{GP}$ & 4.12  & 2.77  & 2.49  \\ \cmidrule{2-2}
			& $d^\prime_{IP}$ & 0.26  & 0.41  & 0.50  \\ \midrule
			\multirow{3}{*}{MIX-D}     & $d^\prime_{GI}$ & 63.71   & 63.71   & 63.71   \\ \cmidrule{2-2}
			& $d^\prime_{GP}$ & 3.99  & 1.86  & 1.34  \\ \cmidrule{2-2}
			& $d^\prime_{IP}$ & 0.21  & 0.82  & 1.22  \\ \bottomrule
		\end{tabular}
	\end{table}
	
	Irreversibility of the CFVNet-based recognition system is analyzed through both
	the BWR template protection method and DCNN. In the BWR method, block remapping is
	reversible in the ideal case. This can be done by determining whether the block
	has been set to the correct position based on an indicator, and such a task is
	essentially like solving a square jigsaw puzzle. If we consider a brute-force
	attack against remapping that tries to rearrange them, the computational
	complexity is $o(n!)$. However, in the deep convolution of DCNN, most of the
	extracted feature maps are abstract semantic information, and the links between
	feature block boundaries are suppressed during DCNN feature extraction. This makes
	rearranging blocks by comparing the boundary similarity more difficult, with a
	computational complexity much larger than $o(n!)$. In addition, random warping is
	applied to the features within blocks and the available block boundary linkage
	distributions are further altered, further enhancing irreversibility of the
	recognition system. The resampling rate
	$r$ in the transformation sub-module sets an upper bound on the reversible
	biometric information. The applied mesh warping can be considered irreversible
	because of the applied interpolation strategy, which leads to information loss.
	Therefore, it is impossible to fully recover the original data even if the
	distortion parameters are known. The deconvolution recovers the original
	biometrics by constructing valid training data, which is easily achieved in
	standard DCNN, whereas in protected CFVNet, the user and his/her original
	biometrics are not linked, and thus no valid training data can be constructed to
	train the deconvolution. When processing biometric features, DCNN-based
	recognition systems are deformed and compressed several times by multiple convolutional and nonlinear transformation layers (activation functions such as ReLU, pooling operations, etc.), which makes the input data be transformed into highly abstract feature representations. The nonlinear transformations also introduce information loss. Moreover, the DCNN-based recognition system only saves the model-trained weight parameters for recognition and does not additionally save any information about the biometric features, making it extremely difficult to recover the original biometric features.
	
	\begin{table}[t]
		\centering
		\renewcommand\arraystretch{0.8}
		\setlength\tabcolsep{1.8pt}
		\caption{Recognition Performance of CFVNet-based Recognition System on Normal and Stolen Scenarios.}
		\label{tb9}
		\begin{tabular}{ccccccccccc}
			\toprule
			\multirow{2}{*}{Scheme} & \multicolumn{2}{c}{HKPU-FV}                         & \multicolumn{2}{c}{SDUMLA-FV}                       & \multicolumn{2}{c}{UTFVP}                           & \multicolumn{2}{c}{FV-USM}                          & \multicolumn{2}{c}{MIX-D}        \\ \cmidrule{2-11} 
			& \multicolumn{1}{c|}{ACC} & \multicolumn{1}{c|}{EER} & \multicolumn{1}{c|}{ACC} & \multicolumn{1}{c|}{EER} & \multicolumn{1}{c|}{ACC} & \multicolumn{1}{c|}{EER} & \multicolumn{1}{c|}{ACC} & \multicolumn{1}{c|}{EER} & \multicolumn{1}{c|}{ACC} & EER   \\ \midrule
			1                       & 0.19                     & 49.61                    & 0.16                     & 50.71                    & 0.00                     & 49.23                    & 0.00                     & 50.31                    & 0.00                     & 49.24 \\
			2                       & 13.79                     & 21.07                    & 9.43                     & 23.27                    & 9.44                     & 16.97                    & 5.28                     & 32.51                    & 8.09                     & 14.55 \\
			3                       & 22.03                    & 17.05                    & 15.75                    & 19.36                    &11.39                     & 13.82                    & 7.52                     & 30.49                    & 14.85                    & 10.59  \\ \bottomrule
		\end{tabular}
	\end{table}
	
	We also evaluate the recognition performance of the CFVNet-based
	recognition system in three scenarios, as shown in Table \ref{tb9}. In the normal
	scenarios, the attacker cannot access the recognition system. When the attacker
	steals system information, security of the FVR system is affected.

	\subsection{Comparison With SOTA Methods}
	\label{sec:sota}
	In this section, we compare the proposed FVR system with state-of-the-art
	schemes. Tables \ref{tb10} and \ref{tb11} compare the recognition accuracy and EER,
	respectively,
	of the latest FVR schemes on each dataset.
	The best
	results of the protected and unprotected schemes are shown in bold. As
	mainstream classification and detection models do not make special modifications for the
	FVR task, their recognition accuracies and EERs are not satisfactory.
	For example, MobileNet is hardly sufficient for basic recognition tasks in
	the presence of complex noise (HKPU-FV and SDUMLA-FV) or insufficient training data
	(UTFVP).
	Larger or more complex 
	models,
	such as EfficientNet
	and Vision Transformer, require larger-scale training data or more learning
	epochs. As for the common object detection framework, it usually searches for
	possible objects in the input image. Even though it accomplishes the basic
	recognition task, the recognition confidence of a single object is low, 
	leading to a recognition system with high FAR, FRR and EER, and is unsuitable for
	high-security recognition scenarios. The proposed CFVNet for FVR
	achieves an average recognition accuracy of 99.82\% and an average EER of
	0.01\% on the four
	publicly available datasets.  We also compare the CFVNet-based system's global linkability with existing state-of-the-art
	schemes in Table \ref{tb12}. Combining localization performance analysis, model
	size, inference time and template protection attributes analysis, using CFVNet to design a cancelable FVR system outperforms existing models in terms of efficiency and performance. It can be seen that based on the proposed CFVNet model,
	an efficient, secure and end-to-end FVR scheme can be
	designed for critical applications. The vein feature information during the
	recognition process is highly irreversible, unlinkable and revocable, which
	effectively improved the template protection attributes of FVR system.
	
	\begin{table}[t]
		\caption{Comparison among latest finger vein recognition schemes recognition accuracy (\%) on each dataset.}
		\label{tb10}
		\centering
		\renewcommand\arraystretch{0.9}
		\setlength\tabcolsep{1pt}
		\begin{tabular}{ccccccc}
			\toprule
			& Method               & Year & HKPU-FV      & SDUMLA-FV      & UTFVP          & FV-USM       \\ \midrule
			\multicolumn{1}{c|}{\multirow{14}{*}{\rotatebox{90}{Unprotected}}} & MobileNet            & 2023 & 42.15        & 61.71          & 23.89          & 100          \\
			\multicolumn{1}{c|}{}                              & ResNet50             & 2023 & 97.89        & 91.90          & 61.94          & 100          \\
			\multicolumn{1}{c|}{}                              & EfficientNet-B7      & 2023 & 99.81        & 94.58          & 92.78          & 99.76        \\
			\multicolumn{1}{c|}{}                              & Vision Transformer                  & 2023 & 79.31        & 78.38          & 21.39          & 98.58        \\
			\multicolumn{1}{c|}{}                              & SSD300               & 2023 & 90.04        & 93.40          & 74.44          & 99.80        \\
			\multicolumn{1}{c|}{}                              & Faster RCNN          & 2023 & 97.31        & 95.68          & 77.50          & 98.78        \\
			\multicolumn{1}{c|}{}                              & Das et al. \cite{das2018convolutional}                     & 2018 & 71.11        & 97.47          & 95.56          & 72.97        \\
			\multicolumn{1}{c|}{}                              & Prasad et al. \cite{prasad2023multi}                     & 2018 & 97.81        & 92.37          & 91.97          & 99.88        \\
			\multicolumn{1}{c|}{}                              & PLS-DA \cite{zhang2021joint}              & 2022 & -            & 97.52          & -              & 99.86        \\
			\multicolumn{1}{c|}{}                              & SSP-DBFL \cite{zhao2023neglected}            & 2023 & 99.39        & 99.07          & -              & 99.40        \\
			\multicolumn{1}{c|}{}                              & RGCN \cite{wang2023residual}                & 2023 & -            & 98.04          & 92.22          & 100 \\
			\multicolumn{1}{c|}{}                              & CFVNet-L (Ours)        & 2023 & 98.85        & 94.97          & 78.33          & 100          \\
			\multicolumn{1}{c|}{}                              & CFVNet-S (Ours)        & 2023 & 98.66        & 95.13          & 77.78          & 99.80        \\ \midrule
			\multicolumn{1}{c|}{\multirow{5}{*}{\rotatebox{90}{Protected}}}    & BDD-ML-ELM \cite{yang2019securing}          & 2018 & -            & 93.09          & 98.61          & -            \\
			\multicolumn{1}{c|}{}                              & Ren et al. \cite{ren2021finger}                    & 2022 & 98.40        & 96.70          & -              & 99.19        \\
			\multicolumn{1}{c|}{}                              & Agerrahrou et al. \cite{aherrahrou2023novel}    & 2023 & -            & 97.22          & -              & 99.19        \\
			\multicolumn{1}{c|}{}                              & \textbf{CFVNet (Ours)} & 2024 & \textbf{100} & \textbf{99.84} & \textbf{99.44} & \textbf{100} \\ \bottomrule
		\end{tabular}
	\end{table}
	
	\begin{table}[t]
		\centering
		\renewcommand\arraystretch{0.9}
		\setlength\tabcolsep{1pt}
		\caption{Comparison among latest finger vein recognition schemes EER (\%) on each dataset.}
		\label{tb11}
		\begin{tabular}{ccccccc}
			\toprule
			& Method               & Year & HKPU-FV       & SDUMLA-FV     & UTFVP         & FV-USM       \\ \midrule
			\multicolumn{1}{c|}{\multirow{13}{*}{\rotatebox{90}{Unprotected}}} & MobileNet            & 2023 & 7.26          & 4.44          & 11.04         & 0.00         \\
			\multicolumn{1}{c|}{}                              & ResNet50             & 2023 & 0.10          & 1.50          & 5.56          & 0.00         \\
			\multicolumn{1}{c|}{}                              & EfficientNet-B7      & 2023 & 0.002         & 0.71          & 0.69          & 0.0004       \\
			\multicolumn{1}{c|}{}                              & Vision Transformer   & 2023 & 2.90          & 5.11          & 15.08         & 0.41         \\
			\multicolumn{1}{c|}{}                              & SSD300               & 2023 & 49.06         & 45.97         & 49.70         & 47.96        \\
			\multicolumn{1}{c|}{}                              & Faster RCNN          & 2023 & 45.37         & 42.64         & 44.69         & 42.20        \\
			\multicolumn{1}{c|}{}                              & Kauba et al. \cite{kauba2022towards}        & 2018 & -             & 3.68          & 0.36          & -            \\
			\multicolumn{1}{c|}{}                              & PLS-DA \cite{zhang2021joint}               & 2018 & 1.17          & 2.15          & -             & 0.15         \\
			\multicolumn{1}{c|}{}                              & FVFSNet \cite{10023509}             & 2022 & 0.81          & 1.10          & 2.08          & 0.20         \\
			\multicolumn{1}{c|}{}                              & FV-DGNN \cite{10201897}             & 2023 & 1.43          & 0.67          & -             & -            \\
			\multicolumn{1}{c|}{}                              & CFVNet-L (Ours)        & 2023 & 0.50          & 0.81          & 4.14          & 0.0002       \\
			\multicolumn{1}{c|}{}                              & CFVNet-S (Ours)        & 2023 & 0.15          & 0.55          & 2.59          & 0.00         \\ \midrule
			\multicolumn{1}{c|}{\multirow{11}{*}{\rotatebox{90}{Protected}}}   & BDD-ML-ELM \cite{yang2019securing}           & 2018 & -             & 7.04          & -             & -            \\
			\multicolumn{1}{c|}{}                              & Shao et al. \cite{shao2022template}          & 2022 & 2.68          & -             & -             & 1.72         \\
			\multicolumn{1}{c|}{}                              & Zhang et al. \cite{zhang2020finger}         & 2020 & 0.88          & 0.61          & -             & -            \\
			\multicolumn{1}{c|}{}                              & Block Remapping \cite{kauba2022towards}      & 2022 & -             & 2.16          & 1.81          & 0.00         \\
			\multicolumn{1}{c|}{}                              & Block Warping \cite{kauba2022towards}        & 2022 & -             & 3.50          & 0.32          & 0.15         \\
			\multicolumn{1}{c|}{}                              & Bloom Filters \cite{kauba2022towards}        & 2022 & -             & 11.70         & 2.23          & -            \\
			\multicolumn{1}{c|}{}                              & Ren et al. \cite{ren2021finger}                     & 2022 & 0.31          & 2.14          & -             & 0.21         \\
			\multicolumn{1}{c|}{}                              & GRP-DHI \cite{choudhary2022multi}              & 2022 & 0.38          & 0.61          & -             & -            \\
			\multicolumn{1}{c|}{}                              & GRP-IoM \cite{choudhary2022multi}             & 2022 & -             & 4.20          & 1.54          & -         \\
			\multicolumn{1}{c|}{}                              & Agerrahrou et al. \cite{aherrahrou2023novel}   & 2023 & -             & 1.11          & -             & 1.33         \\
			\multicolumn{1}{c|}{}                              & \textbf{CFVNet (Ours)} & 2024 & \textbf{0.00} & \textbf{0.04} & \textbf{0.01} & \textbf{0.00} \\ \bottomrule
		\end{tabular}
	\end{table}

	\begin{table}[]
		\centering
		\caption{Comparison among latest finger vein template protection schemes global linkability $D_\leftrightarrow^{sys}$ on each dataset.}
		\label{tb12}
		\renewcommand\arraystretch{1}
		\setlength\tabcolsep{1.5pt}
		\begin{tabular}{cccccc}
			\toprule
			Method            & Year                     & HKPU-FV        & SDUMLA-FV      & UTFVP                & FV-USM               \\ \midrule
			ARH \cite{kirchgasser2020finger}              & 2020                     & -              & -              & 0.052                & -                    \\
			Block Remapping \cite{kauba2022towards}   & 2022                     & -              & 0.047          & \textbf{0.016}                & -                    \\
			Block Warping \cite{kauba2022towards}    & 2022                     & -              & 0.462          & 0.435                & -                    \\
			Bloom Filters \cite{kauba2022towards}    & 2022                     & -              & \textbf{0.019}          & 0.027                & -                    \\
			GRP-DHI \cite{hu2022dual}          & \multicolumn{1}{l}{2022} & \textbf{0.020}           & 0.040           & \multicolumn{1}{l}{} & \multicolumn{1}{l}{} \\
			Aherrahrou et al. \cite{aherrahrou2023novel} & 2023                     & 0.030          & -              & -                    & \uline{0.027}                \\
			CFVNet (Ours)           & 2024                     & \uline{0.028} & \uline{0.026} & \uline{0.022}       & \textbf{0.025}       \\ \bottomrule
		\end{tabular}
	\end{table}
	
	\section{Conclusion}
	\label{sec:conclusion}
	
	Different from previous systems, we integrate preprocessing and template
	protection into a integrated deep learning model. We propose an end-to-end cancelable finger vein network. It can be used to design an efficient and
	secure finger vein recognition system. It 
	includes a plug-and-play BWR-ROIAlign unit, which consists of three main
	sub-modules: localization, compression and transformation. The localization module
	realizes automated ROI localization, and benefits from the recognition branch to
	correct the whole model. The parameters of the localization branch is
	optimized towards the smallest difference in the same instances to achieve 
	optimal feature alignment, which is an effective feature soft alignment scheme.
	The compression module effectively removes unstable noise information through ROI
	position coordinates, and uses the proposed De-R Conv to losslessly and effectively
	remove a large number of redundant feature channels (reducing the number of
	parameters in the model by nearly a factor of $4/5$). The transformation module uses the
	proposed BWR template protection method to introduce unlinkability,
	irreversibility and revocability to the system, which greatly improves the
	recognition system's template protection attributes. We performed experiments on four
	public datasets to study the performance and template protection attributes of the CFVNet-based system.
	The average accuracy, average EERs and average $D_\leftrightarrow^{sys}$ on the
	four datasets can reach 99.82\%, 0.01\% and 0.025, respectively, and achieves competitive
	performance compared with the current state-of-the-art.
	\normalem
	\bibliographystyle{IEEEtran}
	\bibliography{ref}
\end{document}